\documentclass[runningheads]{llncs}
\usepackage{graphicx}
\usepackage{tikz}
\usepackage{comment}
\usepackage{amsmath,amssymb} % define this before the line numbering.
\usepackage{color}
\usepackage{epsfig}
\usepackage{amsmath}
\usepackage{tabularx}
\usepackage{booktabs}
\usepackage{multirow}
\usepackage{subfigure}
\usepackage{bm}
\usepackage{algorithm,algpseudocode}
\usepackage{xcolor,colortbl}
\usepackage[colorlinks, citecolor=green]{hyperref}
\usepackage{breakcites}
\usepackage{threeparttable}
\usepackage{wrapfig,lipsum}
\usepackage{marvosym}
\usepackage{makecell}

\usepackage[accsupp]{axessibility}  % Improves PDF readability for those with disabilities.

\newcommand{\degree}{^\circ}

\definecolor{baselinecolor}{gray}{.9}

\begin{document}
% \linenumbers
\pagestyle{headings}
\mainmatter

\makeatletter
\renewcommand*{\@fnsymbol}[1]{\ensuremath{\ifcase#1\or *\or \textrm{\Letter} \or \dagger \or \ddagger\or
   \mathsection\or \mathparagraph\or \|\or **\or \dagger\dagger
   \or \ddagger\ddagger \else\@ctrerr\fi}}
\makeatother

\title{DODA: Data-oriented Sim-to-Real Domain Adaptation for 3D Semantic Segmentation} % Replace with your title

% \begin{comment}
\titlerunning{DODA}
\author{Runyu Ding\inst{1}\thanks{equal contribution} \and
Jihan Yang\inst{1*} \and Li Jiang\inst{2} \and Xiaojuan Qi\inst{1}\thanks{corresponding author}}
\authorrunning{Ding et al.}
\institute{The University of Hong Kong \and MPI for Informatics\\
\email{\{ryding,jhyang,xjqi\}@eee.hku.hk}~~~~~\email{lijiang@mpi-inf.mpg.de}}

\maketitle

\begin{abstract}
  Deep learning approaches achieve prominent success in 3D semantic segmentation. However, collecting densely annotated real-world 3D datasets is extremely time-consuming and expensive. Training models on synthetic data and generalizing on real-world scenarios becomes an appealing alternative, but unfortunately suffers from notorious domain shifts. In this work, we propose a \textbf{D}ata-\textbf{O}riented \textbf{D}omain \textbf{A}daptation (DODA) framework to mitigate pattern and context gaps caused by different sensing mechanisms and layout placements across domains. Our DODA encompasses virtual scan simulation to imitate real-world point cloud patterns and tail-aware cuboid mixing to alleviate the interior context gap with a cuboid-based intermediate domain. The first unsupervised sim-to-real adaptation benchmark on 3D indoor semantic segmentation is also built on 3D-FRONT, ScanNet and S3DIS along with 8 popular Unsupervised Domain Adaptation (UDA) methods. Our DODA surpasses existing UDA approaches by over 13\% on both 3D-FRONT $\rightarrow$ ScanNet and 3D-FRONT $\rightarrow$ S3DIS.
  Code is available at \href{https://github.com/CVMI-Lab/DODA}{https://github.com/CVMI-Lab/DODA}.
\keywords{Domain Adaptation, 3D Semantic Segmentation}
\end{abstract}

%%%%%%%%%%%%%%%%%%%%%%%%%%
%  INTRO
%%%%%%%%%%%%%%%%%%%%%%%%%%
\section{Introduction}\label{sec:intro}
3D semantic segmentation is a fundamental perception task receiving incredible attention from both industry and academia due to its wide applications in robotics, augmented reality, and human-computer interaction, to name a few. Data hungry deep learning approaches have attained remarkable success for 3D semantic segmentation~\cite{qi2017pointnet++,lahoud20193d,thomas2019kpconv,wu2019pointconv,hu2020randla,xu2021paconv}. 
Nevertheless, harvesting a large amount of annotated data is expensive and time-consuming~\cite{armeni20163d,dai2017scannet}.

\begin{figure}[t]
	\centering
	\includegraphics[width=\linewidth]{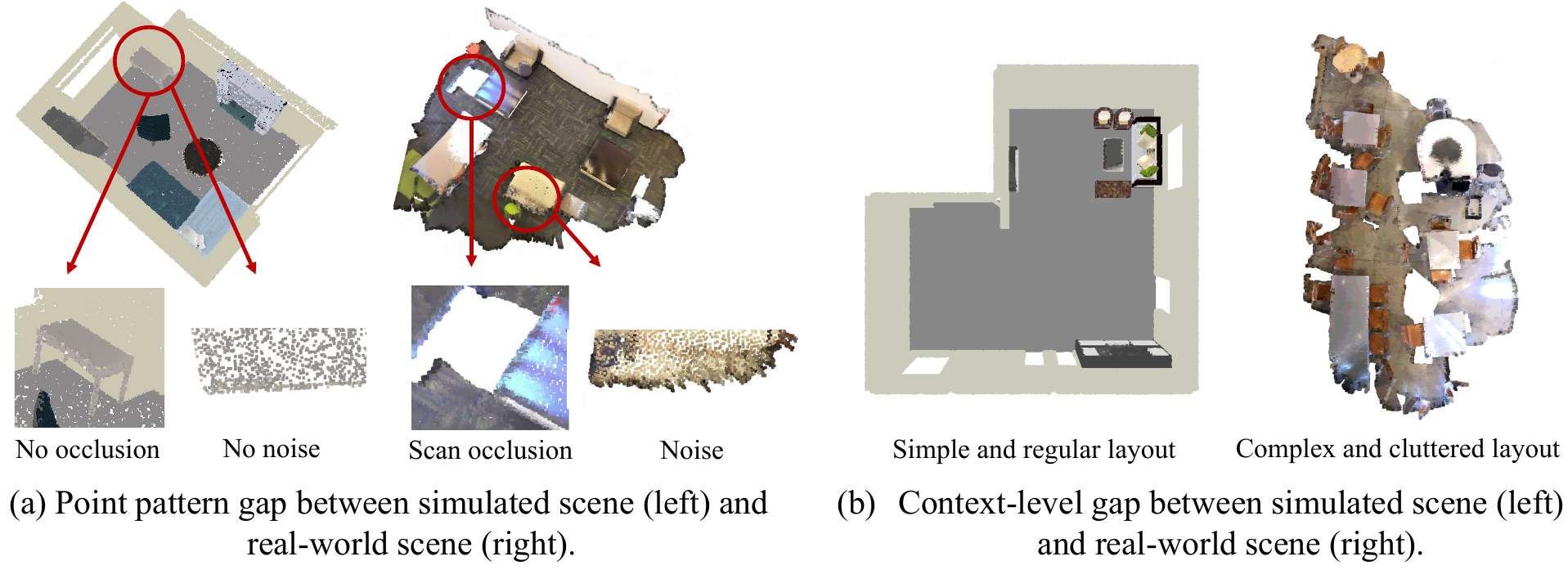}
	\setlength{\abovecaptionskip}{-0.3cm}
    \setlength{\belowcaptionskip}{-0.5cm}
	% captionsetup{font={small}}
	\caption{The domain gaps between simulated scenes from 3D-FRONT~\cite{fu20213d} and real-world scenes from ScanNet~\cite{dai2017scannet}. (a): The point pattern gap. The simulated scene is perfect without occlusions or noise, while the real-world scene inevitably contains scan occlusion and noise patterns such as rough surfaces. (b): The context gap. While the simulated scene applies simple layout with regularly placed objects, the real scene is complex with cluttered interiors.}
    \label{fig:sim_real_gap}
\end{figure}

An appealing avenue to overcome such data scarcity is to leverage simulation data where both data and labels can be obtained for free.
Simulated datasets can be arbitrarily large, easily adapted to different label spaces and customized for various usages~\cite{handa2016understanding,song2017semantic,fu20213d,li2020openrooms,zheng2020structured3d,kar2019meta}.
However, due to notorious domain gaps in point patterns and context (see Fig.~\ref{fig:sim_real_gap}), models trained on simulated scenes suffer drastic performance degradation when generalized to real-world scenarios. 
This motivates us to study sim-to-real unsupervised domain adaptation (UDA), leveraging labeled source data (simulation) and unlabeled target data (real) for effectively adapting knowledge across domains.

Recent efforts on 3D domain adaptation for outdoor scene parsing have obtained considerable progress~\cite{jaritz2020xmuda,wu2019squeezesegv2,zhao2020epointda,kong2021conda}. However, they often adopt LiDAR-specific range image format, not applicable for indoor scenarios with scenes constructed by RGB-D sequences.
Besides, such outdoor attempts could be sub-optimal in addressing the indoor domain gaps raised from different scene construction processes. Further, indoor scenes have more sophisticated interior context than outdoor, which makes the context gap a more essential issue in indoor settings.
Here, we explore sim-to-real UDA in the 3D indoor scenario which is  challenging and largely under explored. 

\noindent
\textit{\textbf{Challenges.}}~
Our empirical studies on sim-to-real adaptation demonstrate two unique challenges in this setting: the \textit{point pattern gap} owing to different sensing mechanisms, and the \textit{context gap} due to dissimilar semantic layouts.
As shown in Fig.~\ref{fig:sim_real_gap} (a), simulated scenes tend to contain complete objects as well as smooth surfaces, while real scenes include inevitable scan occlusions and noise patterns during reconstructing point clouds from RGB-D videos captured by depth cameras~\cite{dai2017scannet,armeni20163d}. Also, even professionally designed layouts in simulated scenes are much simpler and more regular than real layouts as illustrated in Fig.~\ref{fig:sim_real_gap} (b).

To tackle the above domain gaps, we develop a holistic two stage pipeline DODA with a pretrain and a self-training stage, which is widely proved to be effective in UDA settings~\cite{saito2017asymmetric,zou2018unsupervised,yang2021st3d++}. As the root of the challenges lies in ``data'', we thus design two data-oriented modules which are shown to dramatically reduce domain gaps without incurring any computational costs during inference. Specifically, we develop Virtual Scan Simulation (VSS) to mimic occlusion and noise patterns that occur during the construction of real scenes. Such pattern imitation yields a more transferable model to real-world data. 
Afterwards, to adapt the model to target domain, we design Tail-aware Cuboid Mixing (TACM) for boosting self-training. While source supervision is utilized to stabilize gradients with clean labels in self-training, it unfortunately introduces context bias. Thus, we propose TACM to create an intermediate domain by splitting, permuting, mixing and re-sampling source and target cuboids, which explicitly mitigates the context gap through breaking and rectifying source bias with target pseudo-labeled data, and simultaneously eases long-tail issue by oversampling tail cuboids.

To the best of our knowledge, we are the first to explore unsupervised domain adaptation on 3D indoor semantic segmentation. To verify the effectiveness of our DODA, we construct the first 3D indoor sim-to-real UDA benchmark on a simulated dataset 3D-FRONT~\cite{fu20213d} and two widely used real-world scene understanding datasets ScanNet~\cite{dai2017scannet} and S3DIS~\cite{armeni20163d} along with 8 popular UDA methods with task-specific modifications as our baselines. Experimental results show that DODA obtains 22\% and 19\% performance gains in terms of mIoU compared to source only model on 3D-FRONT $\rightarrow$ ScanNet and 3D-FRONT $\rightarrow$ S3DIS respectively. Even compared to existing UDA methods, over 13\% improvement is still achieved. It is also noteworthy that the proposed VSS can lift previous UDA methods by a large margin (8\% $\sim$ 14\%) as a plug-and-play data augmentation, and TACM further facilitates real-world cross-site adaptation tasks with 4\% $\sim$ 5\% improvements.  

%%%%%%%%%%%%%%%%%%%%%%%%%%
%  RELATED WORK
%%%%%%%%%%%%%%%%%%%%%%%%%%
\section{Related Work}
\noindent
\textbf{3D Indoor Semantic Segmentation} focuses on obtaining point-wise category predictions from point clouds, which is a fundamental while challenging task due to the irregularity and sparsity of 3D point clouds. Some previous works~\cite{maturana2015voxnet,song2017semantic} feed 3D grids constructed from point clouds into 3D convolutional neural networks. Some approaches~\cite{graham20183d,choy20194d} further employ sparse convolution~\cite{graham2017submanifold} to leverage the sparsity of 3D voxel representation to accelerate computation. Another line of works~\cite{qi2017pointnet,qi2017pointnet++,jiang2019hierarchical,wu2019pointconv,zhao2019pointweb} directly extract feature embeddings from raw point clouds with hierarchical feature aggregation schemes. Recent methods~\cite{thomas2019kpconv,xu2021paconv} assign position-related kernel functions on local point areas to perform dynamic convolutions. Additionally, graph-based works ~\cite{simonovsky2017dynamic,landrieu2018large,wang2019dynamic} adopt graph convolutions to mimic point cloud structure for point representation learning. 
Although the above methods achieve prominent performance on various indoor scene datasets, they require large-scale human-annotated datasets which we aim to address using simulation data. Our experimental investigation is built upon the sparse-convolution-based U-Net~\cite{graham20183d,choy20194d} due to its high performance.

\noindent
\textbf{Unsupervised Domain Adaptation} aims at adapting models obtained from annotated source data towards unlabeled target samples. The annotation efficiency of UDA and existing data-hungry deep neural networks make it receive great attention from the computer vision community.
Some previous works~\cite{long2015learning,long2017deep} attempt to learn domain-invariant representations by minimizing maximum mean discrepancy~\cite{borgwardt2006integrating}. % of embeddings. 
Another line of research leverages adversarial training~\cite{goodfellow2014generative} to align distributions in feature~\cite{ganin2014unsupervised,saito2019strong,hoffman2017cycada}, pixel~\cite{hoffman2016fcns,hoffman2017cycada,gong2019dlow} or output space~\cite{tsai2018learning} across domains.
Adversarial attacks~\cite{goodfellow2014explaining} have also been utilized in \cite{liu2019transferable,yang2020adversarial} to train domain-invariant classifiers.
Recently, Self-training has been investigated in addressing this problem~\cite{saito2017asymmetric} which formulate UDA as a supervised learning problem guided by pseudo-labeled target data and achieves state-of-the-art performance in semantic segmentation~\cite{zou2018unsupervised} and object detection~\cite{khodabandeh2019robust,simrod}.

Lately, with the rising of 3D vision tasks, UDA has also attracted a lot of attention in such 3D tasks as 3D object classification~\cite{qin2019pointdan,achituve2021self}, 3D outdoor semantic segmentation~\cite{wu2019squeezesegv2,jaritz2020xmuda,yi2021complete,Peng_2021_ICCV,kong2021conda} and 3D outdoor object detection~\cite{wang2020train,yang2021st3d,zhang2021srdan,luo2021consistency,yang2021st3d++}. Especially, Wu \etal~\cite{wu2019squeezesegv2} propose intensity rendering, geodesic alignment and domain calibration modules to align sim-to-real gaps of outdoor 3D semantic segmentation datasets. Jaritz \etal~\cite{jaritz2020xmuda} explore multi-modality UDA by leveraging images and point clouds simultaneously. Nevertheless, no previous work studies UDA on 3D indoor scenes. The unique point pattern gap and the context gap also render 3D outdoor UDA approaches not readily applicable to indoor scenarios. Hence, in this work, we make the first attempt on UDA for 3D indoor semantic segmentation. Particularly, we focus on the most practical and challenging scenario -- simulation to real adaptation.

%%%%\vspace{3pt}
\noindent
\textbf{Data Augmentation for UDA} has also been investigated to remedy data-level gaps across domains. Data augmentation techniques have been widely employed to construct an intermediate domain ~\cite{simrod,kong2021conda,gong2019dlow} to benefit optimization and facilitate gradual domain adaptation.
However, they mainly focus on image-like input formats, which is not suitable for sparse and irregular raw 3D point clouds. Different from existing works, we build a holistic pipeline with two data-oriented modules on two stages to manipulate raw point clouds for mimicking target point cloud patterns and creating a cuboid-based intermediate domain.

%%%%%%%%%%%%%%%%%%%%
% METHOD
%%%%%%%%%%%%%%%%%%%%
\section{Method}
\subsection{Overview}
In this work, we aim at adapting a 3D semantic scene parsing model trained on a source domain $\mathcal{D}_s = \{(P^s_i, Y^s_i)\}_{i=1}^{N_s}$ of $N_s$ samples to an unlabeled target domain $\mathcal{D}_t = \{P^t_i\}_{i=1}^{N_t}$ of $N_t$ samples. $P$ and $Y$ represent the point cloud and the point-wise semantic labels respectively.

\begin{figure*}[t]
	\centering
	\includegraphics[width=\linewidth]{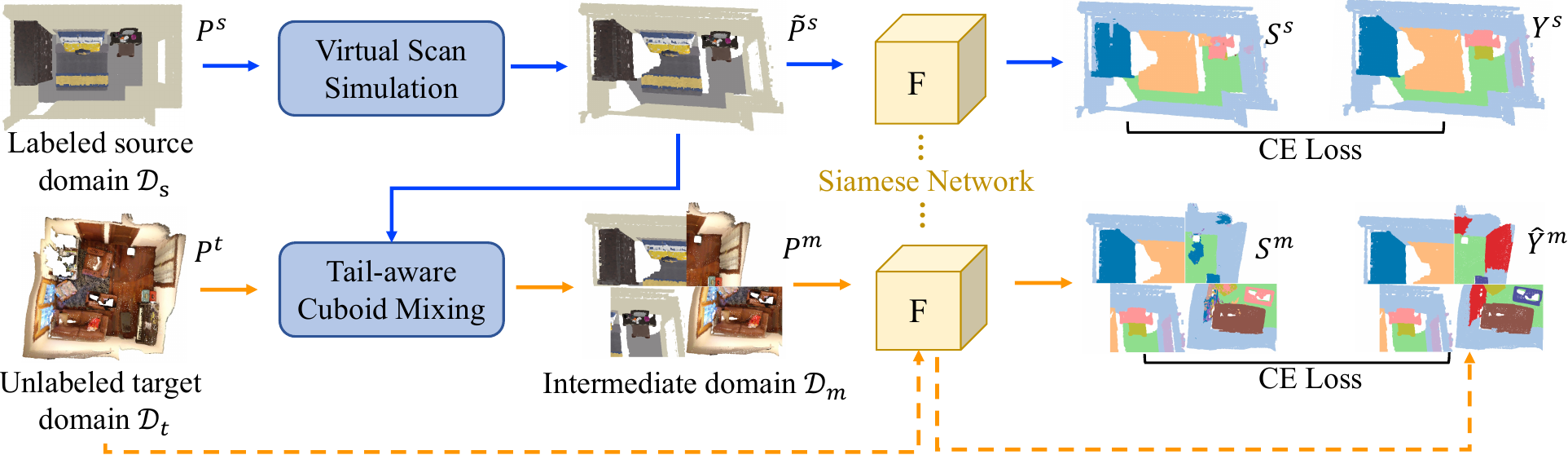}
	\setlength{\abovecaptionskip}{-0.3cm}
    \setlength{\belowcaptionskip}{-0.3cm}
	\caption{Our DODA framework consists of two data-oriented modules: Virtual Scan Simulation (VSS) and Tail-aware Cuboid Mixing (TACM). VSS mimics real-world data patterns and TACM constructs an intermediate domain through mixing source and target cuboids. $P$ denotes the point cloud; $Y$ denotes the semantic labels and $\hat{Y}$ denotes the pseudo labels. The superscripts $s$, $t$ and $m$ stand for source, target and intermediate domain, respectively. The {\color{blue} blue line} denotes source training flow; the {\color{orange} orange line} denotes target training flow and the {\color{orange} orange dotted line} denotes target pseudo label generation procedure. Best viewed in color.}
    \label{fig:framework}
\end{figure*}

In this section, we present DODA, a data-oriented domain adaptation framework to simultaneously close pattern and context gaps by imitating target patterns as well as breaking source bias with the generated intermediate domain. Specifically, as shown in Fig.~\ref{fig:framework}, DODA begins with pretraining the 3D scene parsing model $F$ on labeled source data with our proposed virtual scan simulation module for better generalization. VSS puts virtual cameras on the feasible regions in source scenes to simulate occlusion patterns, and jitters source points to imitate sensing and reconstruction noise in the real scenes. The pseudo labels are then generated with the pretrained model. In the self-training stage, we develop tail-aware cuboid mixing to build an intermediate domain between source and target, which is constructed by splitting and mixing cuboids from both domains. Besides, cuboids including high percentage tail classes are over-sampled to overcome the class imbalance issue during learning with pseudo labeled data. Elaborations of our tailored VSS and TACM are presented in the following parts. 

\subsection{Virtual Scan Simulation}\label{sec:method_vss}
DODA starts from training a 3D scene parsing network on labeled source data, to provide pseudo labels on the target domain in the next self-training stage. Hence, a model with a good generalization ability is highly desirable. 
As analyzed in Sec.~\ref{sec:intro}, different scene construction procedures cause point pattern gaps across domains, significantly hindering the transferability of source-trained models.
Specifically, we find that the missing of occlusion patterns and sensing or reconstruction noise in simulation scenes raises huge negative transfer during the adaptation, which cannot be readily addressed by previous UDA methods (see Sec.~\ref{sec:exp}). This is potentially caused by the fact that models trained on clean source data are incapable of extracting useful features to handle real-world challenging scenarios with ubiquitous occlusions and noise.
To this end, we propose a plug-and-play data augmentation technique, namely virtual scan simulation, to imitate camera scanning procedure for augmenting the simulation data.

VSS includes two parts: the occlusion simulation that puts virtual cameras in feasible regions of simulated scenes to imitate occlusions in the scanning process, and the noise simulation that randomly jitters point patterns to mimic sensing or reconstruction errors,  through which the pattern gaps are largely bridged.

\begin{figure*}[h]
	\centering
	\includegraphics[width=\linewidth]{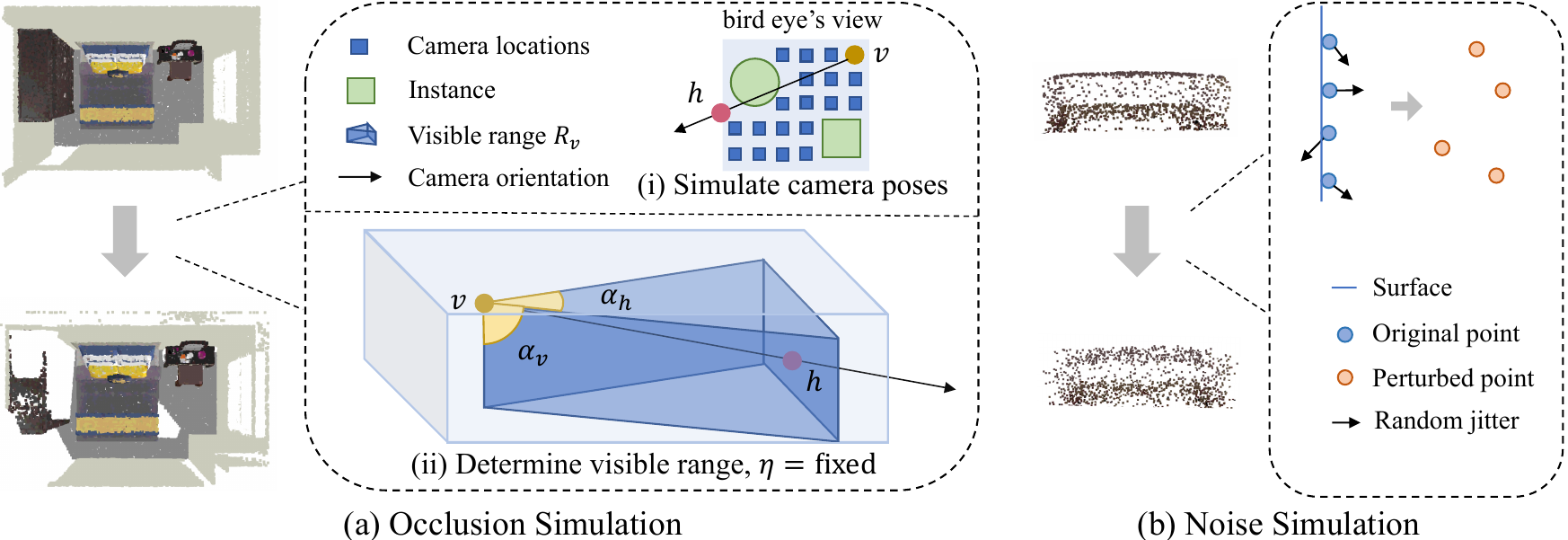}
	\setlength{\abovecaptionskip}{-0.3cm}
    \setlength{\belowcaptionskip}{-0.4cm}
	\caption{Virtual scan simulation. (a): We simulate occlusion patterns by simulating camera poses and determining visible ranges. (b): We simulate noise by randomly jittering points to generate realistic irregular point patterns such as rough surfaces.}
    \label{fig:VSS}
\end{figure*}

\noindent
\textbf{Occlusion Simulation.}~
Scenes in real-world datasets are reconstructed from RGB-D frame sequences suffering from inevitable occlusions, while simulated scenes contain complete objects without any hidden points.
We attempt to mimic occlusion patterns on the simulation data by simulating the real-world data acquisition procedures. Specifically, we divide it into the following three steps:

\noindent
\textit{a) Simulate camera poses.} 
To put virtual cameras in a given simulation scene, we need to determine camera poses including camera positions and camera orientations. 
First, feasible camera positions where a handheld camera can be placed are determined by checking free space in the simulated environment. We voxelize and project $P^s$ to bird eye's view and remove voxels containing instance or room boundary. The centers of remaining free-space voxels are considered as feasible x-y coordinates for virtual cameras, as shown in Fig.~\ref{fig:VSS} (a) (i). 
For the z axis, we randomly sample the camera height in the top half of the room.

Second, for each camera position $v$, we randomly generate a camera orientation using the direction from the camera position $v$ to a corresponding randomly sampled point of interest $h$ on the wall, as shown in Fig.~\ref{fig:VSS} (a) (i). This ensures that simulated camera orientations are uniformly distributed among all potential directions without being influenced by scene-specific layout bias.

\noindent
\textit{b) Determine visible range.}
Given a virtual camera pose and a simulated 3D scene, we are now able to determine the spatial range that the camera can cover, \ie, $R_v $, which is determined by the camera field of view (FOV) (see Fig.~\ref{fig:VSS} (a) (ii)). To ease the modeling difficulties, we decompose FOV into the horizontal viewing angle $\alpha_h$, the vertical viewing angle $\alpha_v$ and the viewing mode $\eta$ that determine horizontal range, vertical range and the shape of viewing frustum, respectively. For the viewing mode $\eta$, we approximate three versions from simple to sophisticated, namely fixed, parallel and perspective, with details presented in the supplementary materials. As illustrated in Fig.~\ref{fig:VSS} (a) (ii), we show an example of the visible range $R_v$ with random $\alpha_h$ and $\alpha_v$ and $\eta$ in the fixed mode.

\noindent
\textit{c) Determine visible points.} 
After obtaining the visible range $R_v$, we then determine the visibility of each point within $R_v$. Specifically, we convert the point cloud to the camera coordinate and extend \cite{katz2007direct} with spherical projection to filter out occluded points and obtain visible points. By taking the union of visible points from all virtual cameras, we finally obtain the point set $P^s_v$ with occluded points removed.
Till now, we can generate occlusion patterns in simulation scenes by mimicking real-world scanning process and adjust the intensity of occlusion by changing the number of camera positions $n_v$ and FOV configurations to ensure that enough semantic context is covered for model learning.

\noindent
\textbf{Noise Simulation.}~
Besides occlusion patterns, sensing and reconstruction errors are unavoidable when generating 3D point clouds from sensor-captured RGB-D videos, which unfortunately results in non-uniform distributed points and rough surfaces in real-world datasets (See Fig.~\ref{fig:sim_real_gap}  (a)). 
To address this issue, we equip our VSS with another noise simulation module, which injects perturbations to each point as follows:

\begin{equation}
\setlength\abovedisplayskip{3pt}
\setlength\belowdisplayskip{3pt}
\tilde{P}^s = \{p + \Delta p \ | \ p \in P^s_v \},
\end{equation}
where $\Delta_p$ denotes the point perturbation following a uniform distribution ranging from $-\delta_p$ to $\delta_p$,
and $\tilde{P}^s$ is the perturbed simulation point cloud. Though simple, we argue that this module efficiently imitates the noise in terms of non-uniform and irregular points patterns as illustrated in Fig.~\ref{fig:VSS}.

\noindent
\textbf{Model Pretraining on Source Data.}~
By adopting VSS as a data augmentation for simulated data, we train a model with cross-entropy loss as Eq.~\eqref{eq:ce1} following settings in~\cite{jiang2021guided,liu2021one}.
\begin{equation}
\setlength\abovedisplayskip{3pt}
\setlength\belowdisplayskip{3pt}
    \min \mathcal{L}_{pre} = \sum\limits_{i=1}^{N_s} \text{CE}(S^s_i, Y^s_{i}),
   \label{eq:ce1}
\end{equation}
where $\text{CE}(\cdot,\cdot)$ is the cross-entropy loss and $S$ is the predicted semantic scores after performing \textit{softmax} on logits.

\subsection{Tail-aware Cuboid Mixing}\label{sec:method_cm}
\setlength\intextsep{2pt}
\begin{wrapfigure}{r}{0.45\textwidth}
\setlength\abovecaptionskip{2pt}
\centering
\includegraphics[width=0.45\textwidth]{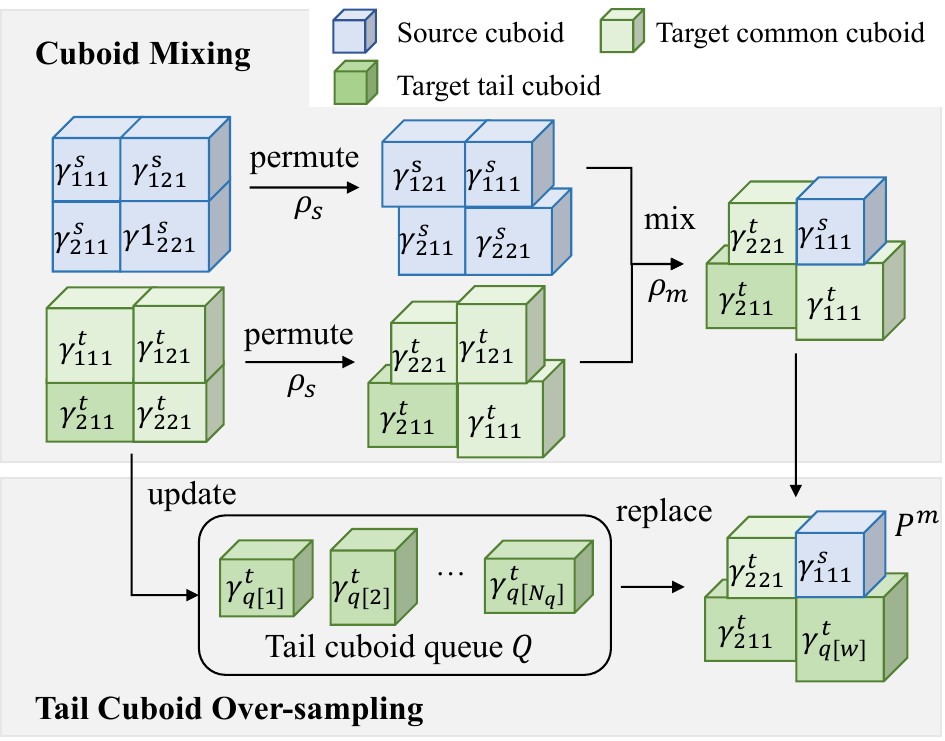}
\caption{An illustration of tail-aware cuboid mixing, which contains cuboid mixing and tail cuboid over-sampling. Notice that for clarity, we take $(n_x,n_y,n_z)=(2,2,1)$ as an example.}
\label{fig:TACM}
\end{wrapfigure}

After obtaining a more transferable scene parsing model with VSS augmentation, we further adopt self-training~\cite{lee2013pseudo,tarvainen2017mean,xie2020self,zou2018unsupervised,zhang2021srdan}, to adapt the model by directly utilizing target pseudo-labeled data for supervision. Since target pseudo label is rather noisy, containing incorrect pseudo labeled data and leading to erroneous supervisions~\cite{yang2021st3d}, we also introduce source supervision to harvest its clean annotations and improve the percentage of correct labels. However, directly utilizing source data unfortunately brings source bias and large
discrepancies in joint optimization. Even though point pattern gaps have already been alleviated with the proposed VSS, the model still suffers from the context gap due to different scene layouts. Fortunately, the availability of target domain data gives us the chance to rectify such context gaps. To this end, we design Tail-aware Cuboid Mixing (TACM) to construct an intermediate domain $\mathcal{D}_m$ that combines source and target cuboid-level patterns (see Fig~\ref{fig:TACM}), which augments and rectifies source layouts with target domain context. Besides, it also decreases the difficulty of simultaneously optimizing source and target domains with huge distribution discrepancies by providing a bridge for adaptation. TACM further moderates the pseudo label class imbalance issue by cuboid-level tail class oversampling. Details on pseudo labeling, cuboid mixing and tail cuboid oversampling are as follows.

\noindent
\textbf{Pseudo Label Generation.}~
To employ self-training after pretraining, we first need to generate pseudo labels $\hat{Y}^t$ for target scenes $P^t$. Similar to previous paradigms~\cite{zou2018unsupervised, xie2020self,yang2021st3d,jiang2021guided}, we obtain pseudo labels via the following equation:
\begin{equation}
\setlength\abovedisplayskip{3pt}
\setlength\belowdisplayskip{3pt}
\hat{Y}^t_{i,j} = \left\{
\begin{array}{lll}
1 &, &\textbf{if}~\max(S^t_i) > T,j=\mathop{\arg\max} S^t_i,  \\
0 & , &\text{otherwise},\\
\end{array}
\right.
\end{equation}
where $\hat{Y}^t_{i}=[\hat{Y}^t_{i,1},\cdots,\hat{Y}^t_{i, c}]$, $c$ is the number of classes and $T$ is the confidence threshold to filter out uncertain predictions.

\noindent
\textbf{Cuboid Mixing.} Here, given labeled source data and pseudo-labeled target data, we carry out the cuboid mixing to construct a new intermediate domain $\mathcal{D}_m$ as shown in Fig.~\ref{fig:framework} and Fig.~\ref{fig:TACM}. For each target scene, we randomly sample a source scene to perform cuboid mixing. We first partition two scenes into several cuboids with varying sizes as the smallest units to mix cuboid as Eq.~\eqref{eq: cuboid}:
\begin{gather}
    P = \{\gamma_{ijk}\},  i \in \{1,  ..., n_x\}, j \in \{1,  ..., n_y\}, k \in \{1, ..., n_z\}, \nonumber\\ 
    \gamma_{ijk}=\{p \mid p~\text{in}~[x_{i-1}, y_{j-1}, z_{k-1}, x_{i}, y_{j}, z_{k}]\},  \label{eq: cuboid}  
\end{gather}
where $\gamma_{ijk}$ denotes a single cuboid; $n_x$, $n_y$ and $n_z$ stand for the number of partitions in $x$, $y$ and $z$ axis, respectively; and each cuboid $\gamma_{ijk}$ is constrained in a six-tuple bounding box $[x_{i-1}, y_{j-1}, z_{k-1}, x_{i}, y_{j}, z_{k}]$ defined by the partition positions $x_i, y_j, z_k$  for corresponding dimensions, respectively.
These partition positions are first initialized as equal-divisions and then injected with randomness to enhance diversities as below: 
\begin{equation}
\setlength\abovedisplayskip{3pt}
\setlength\belowdisplayskip{3pt}
    x_{i}= \begin{cases}
        \frac{i}{n_x}\max p_x+(1-\frac{i}{n_x})\min p_x, &\textbf{if}~i\in\{0, n_x\},\\
        \frac{i}{n_x}\max p_x+(1-\frac{i}{n_x})\min p_x + \Delta \phi, &
        \text{otherwise},
    \end{cases}
\end{equation}
where $\Delta\phi$ is the random perturbation following uniform distribution ranging from $-\delta_\phi$ to $\delta_\phi$. The same formulation is also adopted for $y_j$ and $z_k$. After partitioning, the source and target cuboids are first spatially permuted with a probability $\rho_{s}$ and then randomly mixed with another probability $\rho_{m}$, as depicted in Fig.~\ref{fig:TACM} and Fig.~\ref{fig:framework}.

Though ConDA~\cite{kong2021conda} shares some similarities with our cuboid mixing by mixing source and target, it aims to preserve cross-domain context consistency while ours attempts to mitigate context gaps. Besides, ConDA operates on 2D range images, inapplicable to reconstructed indoor scenes obtained by fusing depth images.
Our cuboid mixing leverages the freedom of the raw 3D representation, \ie~ point cloud, and thus is generalizable to arbitrary 3D scenarios.

\noindent
\textbf{Tail Cuboid Over-sampling.}
Besides embedding target context to source data, our cuboid mixing technique also allows adjusting the category distributions by designing cuboid sampling strategies. Here, as an add-on advantage, we leverage this nice property to alleviate the biased pseudo label problem~\cite{zou2018unsupervised,he2021re,araslanov2021self,liu2021unbiased} in self-training: tail categories only occupy a small percentage of pseudo labeled data. Specifically, we sample cuboids with tail categories more frequently, namely tail cuboid over-sampling, detailed as follows.

We calculate per-class pseudo label ratio $r \in [0, 1]^{c}$ and define $n_r$ least common categories as tail categories. We then define tail cuboid whose pseudo label ratio is higher than the average value $r$ on at least one of $n_r$ tail categories. We construct a tail cuboid queue $Q$ with size $N_q$ to store tail cuboids. Formally, $\gamma^t_{q[w]}$ denotes the $w^{th}$ tail cuboid in $Q$, as shown in Fig.~\ref{fig:TACM}. Notice that through training, $Q$ is dynamically updated with First In, First Out (FIFO) rule since cuboids are randomly split in each iteration as Eq.~\eqref{eq: cuboid}.
In each training iteration, we ensure that at least $u$ tail cuboids are in each mixed scene by sampling cuboids from $Q$ and replacing existing cuboids if needed. 
With such a simple over-sampling strategy, we make the cuboid mixing process tail-aware, and relieve the class imbalance issue in the self-training. Experimental results in Sec.~\ref{sec:ablation} further demonstrate the effectiveness of our tail cuboid over-sampling strategy.

\noindent
\textbf{Self-training with Target and Source data.} In the self-training stage, for data augmentation, VSS is first adopted to augment the source domain data to reduce the pattern gap and then TACM mixes source and target scenes to construct a tail-aware intermediate domain $\mathcal{D}_m=\{P^m\}$ with labels $\hat{Y}^m$ mixed by source ground-truth and target pseudo labels. To alleviate the noisy supervisions from incorrect target pseudo labels, we minimize dense cross-entropy loss on source data $\tilde{P}^s$ and intermediate domain data  $P^m$  as below:
\begin{equation}
    \setlength\abovedisplayskip{3pt}
    \setlength\belowdisplayskip{3pt}
    \min \mathcal{L}_{st} = \sum\limits_{i=1}^{N_t} \text{CE}(S^m_i, \hat{Y}^m_{i}) + \lambda\sum\limits_{i=1}^{N_s} \text{CE}(S^s_i, Y^s_{i}),
\end{equation}
where $\lambda$ denotes the trade-off factor between losses.

\section{Benchmark Setup}
\subsection{Datasets}
\noindent
\textbf{3D-FRONT}~\cite{fu20213d} is a large-scale dataset of synthetic 3D indoor scenes, which contains 18,968 rooms with 13,151 CAD 3D furniture objects from 3D-FUTURE~\cite{fu20213dfuture}. The layouts of rooms are created by professional designers and distinctively span 31 scene categories and 34 object semantic super-classes. 
We randomly select 4995 rooms as training samples and 500 rooms as validation samples after filtering out noisy rooms. Notice that we obtain source point clouds by uniformly sampling points from original mesh with CloudCompare~\cite{girardeau2016cloudcompare} at 1250 surface density (number of points per square units). Comparison between 3D-FRONT and other simulation datasets are detailed in the nsupplemental materials. 

\noindent
\textbf{ScanNet}~\cite{dai2017scannet} is a popular real-world indoor 3D scene understanding dataset, consisting 1,613 real 3D scans with dense semantic annotations (\ie,~1,201 scans for training, 3,12 scans for validation and 100 scans for testing). It provides semantic annotations for 20 categories. 

\noindent
\textbf{S3DIS}~\cite{armeni20163d} is also a well-known real-world indoor 3D point cloud dataset for semantic segmentation. It contains 271 scenes across six areas along with 13 categories with point-wise annotations. Similar to previous works~\cite{li2018pointcnn,qi2017pointnet++}, we use the fifth area as the validation split and other areas as the training split.

\noindent
\textbf{Label Mapping.} Due to different category taxonomy of datasets, we condense 11 categories for 3D-FRONT $\rightarrow$ ScanNet and 3D-FRONT $\rightarrow$ S3DIS settings, individually. Besides, we condense 8 categories for cross-site settings between S3DIS and ScanNet. Please refer to the Suppl. for the detailed taxonomy.

\subsection{UDA Baselines}
As shown in Table~\ref{tab:scannet} and \ref{tab:s3dis}, we reproduce 7 popular 2D UDA methods and 1 3D outdoor method as UDA baselines, encompassing MCD~\cite{saito2018maximum}, AdaptSegNet~\cite{tsai2018learning}, CBST~\cite{zou2018unsupervised}, MinEnt~\cite{vu2019advent}, AdvEnt~\cite{vu2019advent}, Noisy Student~\cite{xie2020self}, APO-DA~\cite{yang2020adversarial} and SqueezeSegV2~\cite{wu2019squeezesegv2}. These UDA baselines cover most existing streams such as adversarial alignment, discrepancy minimization, self-training and entropy guided adaptation. To perform these image-based methods on our setting, we carry out some task-specific modifications, which are detailed in supplemental materials.

%%%%%%%%%%%%%%%%%%%%%%%%%%%%%%%%% EXPERIMENTS
\section{Experiments}\label{sec:exp}
To validate our method, we benchmark DODA and other popular UDA methods with extensive experiments on 3D-FRONT~\cite{fu20213d}, ScanNet~\cite{dai2017scannet} and S3DIS~\cite{armeni20163d}. Moreover, we explore a more challenging setting, from simulated 3D-FRONT~\cite{fu20213d} to RGBD realistic dataset NYU-V2~\cite{Silberman:ECCV12}, presented in the supplementary materials.
To verify the generalizability of VSS and TACM, we further integrate VSS to previous UDA methods and adopt TACM in the real-world cross-site UDA setting.
Note that since textures for some background classes are not provided in 3D-FRONT dataset, we only focus on adaptation using 3D point positions. The implementation details including network and training details are provided in the Suppl.

% \subsection{Main Results}
\noindent
\textbf{Comparison to Other UDA Methods.}
As shown in Table~\ref{tab:scannet} and Table~\ref{tab:s3dis}, compared to source only, DODA largely lifts the adaptation performance in terms of mIoU by around 21\% and 19\% on 3D-FRONT $\rightarrow$ ScanNet and 3D-FRONT $\rightarrow$ S3DIS, respectively.
DODA also shows its superiority over other popular UDA methods, obtaining $14\% \sim 22\%$ performance gain on 3D-FRONT $\rightarrow$ ScanNet and $13\% \sim 19\%$ gain on 3D-FRONT $\rightarrow$ S3DIS.
Even only equipping source only with VSS module, our DODA (only VSS) still outperforms UDA baselines by around $4\% \sim 10\%$, indicating that the pattern gap caused by different sensing mechanisms significantly harms adaptation results while previous methods have not readily addressed it.
Comparing DODA with DODA (w/o TACM), we observe that TACM mainly contributes to the performance of instances such as bed and bookshelf on ScanNet, since cuboid mixing forces model to focus more on local semantic clues and object shapes itself inside cuboids.
It is noteworthy that though DODA yields general improvement around almost all categories adaptation in both pretrain stage and self-training stage, challenging classes such as bed on ScanNet and sofa on S3DIS attain more conspicuous performance lift, demonstrating the predominance of DODA in tackling troublesome categories. However, the effectiveness of all UDA methods for column and beam on S3DIS are not obvious due to their large disparities in data patterns across domains and low appearing frequencies in source domain. To illustrate the reproducibility of our DODA, all results are repeated three times and reported as average performance along with standard variance.

% 3dfront -> scannet, main results
\setlength\intextsep{8pt}
\begin{table*}[h]
    \centering
    \setlength\belowcaptionskip{2pt}
    \caption{Adaptation results of 3D-FRONT $\rightarrow$ ScanNet in terms of mIoU. We indicate the best adaptation result in \textbf{bold}. $\dagger$ denotes pretrain generalization results with VSS}
    \begin{small}
      \scalebox{0.67}{
        \setlength{\tabcolsep}{1.6mm}{
        \begin{tabular}{c|c|ccccccccccc}
            \bottomrule[1pt]
            Method & mIoU & wall & floor & cab. & bed & chair & sofa & table & door & wind. & bksf. & desk \\
            \hline
            Source Only & 29.60 & 60.72 & 82.42 & 04.44 & 12.02 & 61.76 & 22.31 & 38.52 & 05.72 & 05.12 & 19.72 & 12.84 \\
            \hline
            MCD~\cite{saito2018maximum} & 32.27 & 62.86 & 88.70 & 03.81 & 38.50 & 57.51 & 21.48 & 41.67 & 05.78 & 01.29 & 18.81 & 15.69\\
            %\hline
            AdaptSegNet~\cite{tsai2018learning} & 34.51 & 61.81 & 83.90 & 03.64 & 36.06 & 55.05 & 34.26 & 44.21 & 06.59 & 05.54 & 31.87 & 16.64 \\
            %\hline
            CBST~\cite{zou2018unsupervised} & 37.42 & 60.37 & 81.39 & 12.18 & 30.00 & 68.86 & 36.22 & 49.93 & 07.05 & 05.82 & 43.59 & 16.25 \\
            %\hline
            MinEnt~\cite{vu2019advent} & 34.61 & 63.35 & 85.54 & 04.66 & 26.05 & 61.98 & 33.05 & 48.38 & 05.20 & 03.15 & 35.84 & 13.49 \\
            %\hline
            AdvEnt~\cite{vu2019advent} & 32.81 & 64.31 & 79.21 & 04.39 & 35.01 & 61.05 & 24.36 & 41.64 & 05.97 & 01.60 & 29.07 & 14.32 \\
            %\hline
            Noisy student~\cite{xie2020self} & 34.67 & 62.63 & 86.27 & 01.45 & 17.13 & 69.98 & 37.58 & 47.87 & 06.01 & 01.66 & 35.79 & 15.06 \\
            %\hline
            APO-DA~\cite{yang2020adversarial} & 31.73 & 62.84 & 85.43 & 02.77 & 15.08 & 64.24 & 34.41 & 46.41 & 03.94 & 03.59 & 18.88 & 11.41 \\
            SqueezeSegV2~\cite{wu2019squeezesegv2} & 29.77 & 61.85 & 72.74 & 02.50 & 16.89 & 58.79 & 16.81 & 38.19 & 05.08 & 03.24 & 35.68 & 15.72 \\
            % \toprule[0.5pt]
            % \bottomrule[0.5pt]
            \hline
            DODA (only VSS)$^\dagger$ & 40.52$\pm$0.80 & 67.36 & 90.24 & {15.98} & {39.98} & {63.11} & {46.38} & {48.05}
            & {07.63} & {13.98} & {33.17} & {19.86} \\
            DODA (w/o TACM) & 48.13$\pm$0.25 & 72.22 & 93.43 & 24.46 & 56.30 & 70.40 & 53.33 & 56.57 & \textbf{09.44} & 19.97 & 47.05 & 26.25 \\
            DODA & \textbf{51.42$\pm$0.90} & \textbf{72.71} & \textbf{93.86} & \textbf{27.61} & \textbf{64.31} & \textbf{71.64} & \textbf{55.30} & \textbf{58.43} & {08.21} & \textbf{24.95} & \textbf{56.49} & \textbf{32.06} \\
            % \toprule[0.5pt]
            % \bottomrule[0.5pt]
            \hline
            Oracle & 75.19 & 83.39 & 95.11 & 69.62 & 81.15 & 88.95 & 85.11 & 71.63 & 47.67 & 62.74 & 82.63 & 59.05 \\
            \toprule[1pt]
        \end{tabular}}}
    \end{small}
    \label{tab:scannet}
\end{table*}

% 3dfront -> s3dis, main results
\begin{table*}[h]
    \centering
    \setlength{\belowcaptionskip}{0.1cm}
    \caption{ Adaptation results of 3D-FRONT $\rightarrow$ S3DIS in terms of mIoU. We indicate the best adaptation result in \textbf{bold}. $\dagger$ denotes pretrain generalization results with VSS}
    \begin{small}
      \scalebox{0.67}{
        \setlength{\tabcolsep}{1.6mm}{
        \begin{tabular}{c|c|ccccccccccc}
            \bottomrule[1pt]
            Method & mIoU & wall & floor & chair & sofa & table & door & wind. & bkcase. & ceil. & beam & col. \\
            \hline
            Source Only & 36.72 & 67.95 & 88.68 & 57.69 & 04.15 & 38.96 & 06.99 & 00.14 & 44.90 & 94.42 & 00.00 & 00.00\\
            \hline
            % \toprule[0.5pt]
            % \bottomrule[0.5pt]
            MCD~\cite{saito2018maximum} & 36.62 & 64.53 & 92.16 & 54.76 & 13.31 & 46.67 & 8.54 & 00.08 & 28.86 & 93.89 & 00.00 & 00.00 \\
            %\hline
            AdaptSegNet~\cite{tsai2018learning} & 38.14 & 68.14 & 93.17 & 55.14 & 05.31 & 43.14 & 14.67 & 00.33 & 45.75 & 93.88 & 00.00 & 00.00 \\
            %\hline
            CBST~\cite{zou2018unsupervised} & 42.47 & 71.60 & 92.07 & 68.09 & 03.28 & 60.45 & 17.13 & 00.18 & 58.45 & 95.87 & 00.00 & 00.00 \\
            %\hline
            MinEnt~\cite{vu2019advent} & 37.08 & 66.15 & 87.92 & 52.30 & 06.27 & 25.79 & 15.70 & 04.44 & 55.72 & 93.58 & 00.00 & 00.00 \\
            %\hline
            AdvEnt~\cite{vu2019advent} & 37.98 & 66.94 & 91.84 & 57.96 & 02.39 & 46.18 & 15.14 & 00.54 & 44.31 & 92.50 & 00.00 & 00.00 \\
            %\hline
            Noisy student~\cite{xie2020self} & 39.44 & 68.84 & 91.78 & 65.53 & 06.65 & 48.67 & 02.27 & 00.00 & 53.67 & \textbf{96.46} & 00.00 & 00.00 \\
            %\hline
            APO-DA~\cite{yang2020adversarial} & 38.23 & 68.63 & 89.66 & 58.84 & 03.51 & 40.66 & 13.73 & 02.61 & 47.88 & 94.97 & \textbf{00.04} & 00.00 \\
            SqueezeSegV2~\cite{wu2019squeezesegv2} & 36.50 & 65.01 & 89.95 & 54.29 & 06.79 & 45.75 & 10.23 & 01.70 & 32.93 & 94.81 & 00.00 & 00.00\\
            \hline
            % \toprule[0.5pt]
            % \bottomrule[0.5pt]
            DODA (only VSS) $^\dagger$ & 46.85$\pm$0.78 & 70.96 & 96.12 & 68.70 & 25.47 & 58.47 & 17.87 & 27.65 & 54.39 & 95.66 & 00.00 & 00.00\\
            %\hline
            DODA (w/o TACM) & 53.86$\pm$0.49 & 75.75 & 95.14 & 76.12 & 60.11 & 64.07 & 25.24 & 31.75 & \textbf{68.49} & 95.82 & 00.00 & 00.00\\
            DODA & \textbf{55.54$\pm$0.91} & \textbf{76.23} & \textbf{97.17} & \textbf{76.89} & \textbf{63.55} & \textbf{69.04} & \textbf{25.76} & \textbf{38.22} & 68.18 & 95.85& 00.00& 00.00\\
            \hline
            % \toprule[0.5pt]
            % \bottomrule[0.5pt]
            Oracle & 62.29 & 82.82 & 96.95 & 78.16 & 40.37 & 78.56 & 56.91 & 47.90 & 77.10 & 96.29 & 00.41 & 29.69 \\
            \toprule[1pt]
        \end{tabular}}}
    \end{small}
    \label{tab:s3dis}
\end{table*}

\noindent
\textbf{VSS Plug-and-play Results to Other UDA Methods.}
Since VSS works as a data augmentation in our DODA, we argue that it can serve as a plug-and-play module to mimic occlusion and noise patterns on simulation data, and is orthogonal to existing UDA strategies.
As demonstrated in Table~\ref{tab:VSS}, equipped with VSS, current popular UDA approaches consistently surpass their original performance by around $8\%\sim 13\%$. % from 3D-FRONT $\rightarrow$ ScanNet.
It also verifies that previous 2D-based methods fail to close the point pattern gap in 3D indoor scene adaptations, while our VSS can be incorporated into various pipelines to boost performance.

\noindent
\textbf{TACM Results in Cross-site Adaptation.}
Serving as a general module to alleviate domain shifts across domains, we show that TACM can consistently mitigate domain discrepancies on even real-to-real adaptation settings. For cross-site adaptation, scenes collected from different sites or room types also suffer a considerable data distribution gap.
As shown in Table~\ref{tab:TACM}, the domain gaps in real-to-real adaptation tasks are also large when comparing the source only and oracle results. When adopting TACM in the self-training pipelines, they obtain 5.64\% and 3.66\% relative performance boost separately in ScanNet $\rightarrow$ S3DIS and S3DIS $\rightarrow$ ScanNet. These results verify that TACM is general in relieving data gaps, especially the context gap on various 3D scene UDA tasks. 
We provide the cross-site benchmark with more UDA methods in the Suppl.

\begin{table}[h]
    \centering
    \makebox[0pt][c]{\parbox{\textwidth}{%
    \begin{minipage}[h]{0.46\hsize}
        \setlength\belowcaptionskip{0.1cm}
        \caption{UDA results equipped with VSS on 3D-FRONT $\rightarrow$ ScanNet}
        \begin{small}
        \scalebox{0.75}{
        \setlength{\tabcolsep}{2mm}{
        \begin{tabular}{c|c|c|c}
            \bottomrule[1pt]
            \multirow{2}{*}{Method} & \multicolumn{2}{c|}{VSS} & \multirow{2}{*}{Improv.}\\
            \cline{2-3}
            & w/o & w/ \\
            \hline
            MCD~\cite{saito2018maximum} & 32.37 & 40.32 & +7.95\\
            AdaptSegNet~\cite{tsai2018learning} & 34.51 & 45.75 & +11.24\\
            CBST~\cite{zou2018unsupervised}  & 36.30 & 47.70 & +11.40\\
            MinEnt~\cite{vu2019advent} & 34.61 & 43.26 & +8.65\\
            AdvEnt~\cite{vu2019advent} & 32.81 & 42.94 & +10.13\\
            Noisy Student~\cite{xie2020self} & 34.67 & 48.30 & +13.63\\
            APO-DA~\cite{yang2020adversarial} & 31.73 & 43.98 & +12.25\\
            SqueezeSegV2~\cite{wu2019squeezesegv2} & 29.77 & 40.60 & +10.83\\
            \toprule[0.8pt]
        \end{tabular}}}
        \end{small}
        \label{tab:VSS}
    \end{minipage}
    \hfill
    \begin{minipage}[h]{0.52\hsize}
        \setlength\belowcaptionskip{0.1cm}
        \caption{Cross-site adaptation results with TACM}
        \begin{small}
        \scalebox{0.82}{
        \setlength{\tabcolsep}{1.6mm}{
        \begin{tabular}{c|c|c}
            \bottomrule[1pt]
            Task & Method & mIoU \\
            \hline
            \multirow{4}{*}{ScanNet$\rightarrow$ S3DIS} & Source Only & 54.09 \\
            & CBST~\cite{zou2018unsupervised} & 60.13 \\
            % \multirow{2}{*}{+5.64}\\
            & CBST+TACM & 65.52 \\
            & Oracle & 72.51 \\
            \hline
            \multirow{4}{*}{S3DIS $\rightarrow$ ScanNet} & Source Only & 33.48\\ 
            & Noisy Student~\cite{xie2020self} & 44.81\\
            & Noisy Student+TACM & 48.47 \\
            & Oracle & 80.06 \\
            \toprule[0.8pt]
        \end{tabular}}}
        \end{small}
        \label{tab:TACM}
    \end{minipage}
    }}
\end{table}

%%%%%%%%%%%%%%%%%%%%%%%%%%%%%% 
% ABLATIONS
%%%%%%%%%%%%%%%%%%%%%%%%%%%%%%

\section{Ablation Study}\label{sec:ablation}

In this section, we conduct extensive ablation experiments to investigate the individual components of our DODA. All experiments are conducted on 3D-FRONT $\rightarrow$ ScanNet for simplicity. Default settings are marked in \textbf{bold}.

\noindent
\textbf{Component Analysis.}
Here, we investigate the effectiveness of each component and module in our DODA.
As shown in Table~\ref{tab:comp_analysis}, occlusion simulation brings the largest performance gain (around $9.7\%$), indicating that model trained on complete scenes is hard to adapt to scenes with occluded patterns.
Noise simulation further supplements VSS to imitate sensing and reconstruction noise, obtaining about $1.3\%$ boosts. Two sub-modules jointly mimic realistic scenes, largely alleviating the point distribution gap and leading to a more generalizable source only model.
In the self-training stage, VSS also surpasses the baseline by around 13\% due to its efficacy in reducing the point pattern gap and facilitating generating high-quality pseudo labels. 
Cuboid mixing combines cuboid patterns from source and target domains for moderating context-level bias, further boosting the performance by around $2.4\%$. Moreover, cuboid-level tail-class over-sampling yields $0.9\%$ improvement with greater gains on tail classes. For instance, desk on ScanNet achieves 6\% gain (see Suppl.).

\begin{table}[htbp]
	\centering
	\setlength{\belowcaptionskip}{0.1cm}
    \caption{Component Analysis for DODA on 3D-FRONT $\rightarrow$ ScanNet}% with regard  to occlusion and noise simulation for VSS, and cuboid mixing and tail c uboid resampling for TACM.}
	\begin{small}
        \scalebox{0.95}{
    	\setlength{\tabcolsep}{1mm}{
            \begin{tabular}{c|cc|cc|c}
                \bottomrule[1pt]
                \multirow{2}{*}{Baseline} & \multicolumn{2}{c|}{Virtual scan simulation} & \multicolumn{2}{c|}{Tail-aware cuboid mixing} & \multirow{2}{*}{mIoU}\\
                \cline{2-5}
                & Occlusion sim. & Noise sim. & Cuboid mix. & Tail samp. & \\
    			\hline
    			Source Only & &&&& 29.60\\
    		    Source Only & \checkmark &&&& 39.25 (+9.65)\\
    		    Source Only & \checkmark & \checkmark &&&  40.52 (+1.27) \\
    		    \hline
                Noisy Student &  &  &&& 34.67 \\
    		    Noisy Student & \checkmark & \checkmark &&& 48.13 (+13.46)\\
    		    Noisy Student & \checkmark & \checkmark & \checkmark && 50.55 (+2.42)\\
    		    Noisy Student & \checkmark & \checkmark & \checkmark & \checkmark & \textbf{51.42} (+0.87)\\
                \toprule[0.8pt]
    		\end{tabular}}}
	\end{small}
    \label{tab:comp_analysis}
\end{table}

\noindent
\textbf{VSS: Visible Range.} Here, we study the effect of visible range of VSS, which is jointly determined by the horizontal angle $\alpha_h$, vertical angle $\alpha_v$, viewing mode $\eta$ and the number of cameras $n_v$. As shown in Table~\ref{tab:fov}, fewer cameras $n_v=2$ and smaller viewing angle $\alpha_v=45\degree$ draw around 2\% performance degradation with a smaller visible range. And decreasing $\alpha_h$ to 90$\degree$ can also achieve similar performance with $\alpha_h = 180\degree$ with more 
cameras $n_v=8$, demonstrating that enough semantic coverage is a vital factor.
Besides, as for the three viewing modes $\eta$, the simplest fixed mode achieves the highest performance in comparison to parallel and perspective modes.
Even though parallel and perspective are more similar to reality practice, they cannot cover sufficient range with limited cameras, since real-world scenes are reconstructed through hundreds or thousands of view frames. This again demonstrates that large spatial coverage is essential. To trade off between the effectiveness and efficiency of on-the-fly VSS, we use fixed mode with 4 camera positions by default here.

\begin{table}[h]
    \begin{center}
    \makebox[0pt][c]{\parbox{\textwidth}{%
    \begin{minipage}[h]{0.51\hsize}
        \setlength{\belowcaptionskip}{0.1cm}
        \caption{Ablation study of visible range design on 3D-FRONT $\rightarrow$ ScanNet}
	    \begin{small}
	    \scalebox{1.0}{
    	\setlength{\tabcolsep}{1.5mm}{
            \begin{tabular}{c|c|c|c|c}
                \bottomrule[1pt]
                $\alpha_h$ & $\alpha_v$ & $\eta$ & $n_v$ & mIoU\\
    			\hline
    			$180\degree$ &  $90\degree$ & fixed & 2 & 38.80 \\
                $180\degree$ & $90\degree$ & fixed & 4 & \textbf{40.52}\\
                $90\degree$ & $90\degree$ & fixed & 8 & 40.30  \\
                \hline
    			$180\degree$ & $90\degree$ & parallel & 4 & 39.08 \\
    		    $180\degree$ & $90\degree$ & perspective & 4 & 39.04  \\
    		    $180\degree$ & $45\degree$ & perspective & 4 & 36.64  \\
                \toprule[0.8pt]
    		\end{tabular}}}
        \end{small}
        \label{tab:fov}
    \end{minipage}
    \hfill
    \begin{minipage}[h]{0.44\hsize}
        \setlength{\belowcaptionskip}{0.1cm}
        \caption{Ablation study of cuboid partitions on 3D-FRONT $\rightarrow$ ScanNet % Spatial partitions $(n_x,n_y,n_z)$ denotes the number of partitions in corresponding dimensions.
        }
        \begin{small}
        \scalebox{0.88}{
        \setlength{\tabcolsep}{1.7mm}{
            \begin{tabular}{c|c|c}
                \bottomrule[1pt]
                ($n_x$, $n_y$, $n_z$) & \# cuboid & mIoU\\
                \hline
                (1, 1, 1) & 1 & 48.10\\
    			\hline
    			(2, 1, 1) & 2 & 50.00\\
                % (3, 1, 1) & 3 & 50.01 \\
    		    (2, 2, 1) & 4 & \textbf{50.55} \\
                (3, 2, 1) & 6 & 50.57 \\
                (3, 3, 1) & 9 & 50.02 \\
                \hline
                (1, 1, 2) & 2 & 49.49 \\
                \hline
                (2, 1, 2) & 4 & 49.48 \\
                % (2, 2, 2) & 8 & 49.50 \\
                \toprule[0.8pt]
    		\end{tabular}}}
        \end{small}
        \label{tab:cuboid}
    \end{minipage}
    }}
    \end{center}
\end{table}

\noindent
\textbf{TACM: Cuboid Partition.} We study various cuboid partition manners in Table~\ref{tab:cuboid}. Notice that random rotation along z axis is performed before cuboid partition, so the partition on x or y axes can be treated as identical. While horizontal partitioning yields consistent performance beyond 50\% mIoU, vertical partitioning does not show robust improvements, suggesting the mixing of vertical spatial context is not necessary. 
Simultaneous partitioning on x and y axes also improves performance (\ie~(2,2,1) and (2,3,1)), while too small cuboid size (\ie~(3,3,1)) results in insufficient context cues in each cuboid with a slight decrease in mIoU.

\noindent
\textbf{TACM: Data-mixing Method.}
We compare TACM with other popular data-mixing methods in Table~\ref{tab:intermediate_domain}. Experimental results show the superiority of TACM since it outperforms Mix3D~\cite{nekrasov2021mix3d}, CutMix~\cite{yun2019cutmix} and Copy-Paste~\cite{ghiasi2021simple} by around $2.2\%$ to $2.9\%$. 
TACM effectively alleviates the context gap while preserving local context clues. Mix3D, however, results in large overlapping areas, which is unnatural and causes semantic confusions. CutMix and Copy-Paste only disrupt local areas without enough perturbations of the broader context (see Suppl.).

\noindent
\textbf{TACM: Tail Cuboid Over-sampling with Class-balanced Loss.}
Tail cuboid over-sampling brings significant gains on tail classes as discussed in Sec.~\ref{sec:ablation}. As demonstrated in Table~\ref{tab:tail-class}, the class-balanced lovasz loss~\cite{berman2018lovasz} also boosts performance by considering each category more equally. We highlight that our TACM can also incorporate with other class-balancing methods during optimization since it eases long tail issue on the data-level.

% cuboid partition & domain-mix methods
\begin{table}[h]
    \centering
    \makebox[0pt][c]{\parbox{\textwidth}{
    \begin{minipage}[hbtp]{0.46\hsize}
       \setlength{\belowcaptionskip}{0.1cm}
        \caption{Ablation study of data-mixing methods on 3D-FRONT $\rightarrow$ ScanNet}
        \begin{small}
        \scalebox{0.85}{
        \setlength{\tabcolsep}{7mm}{
        \begin{tabular}{c|c}
            \bottomrule[1pt]
            Method & mIoU\\
            \hline
            Mix3D~\cite{nekrasov2021mix3d} & 48.62 \\
            CutMix~\cite{yun2019cutmix} & 49.19 \\
            Copy-Paste~\cite{ghiasi2021simple} & 48.51 \\
            \hline
            TACM & \textbf{51.42} \\
            \toprule[0.8pt]
        \end{tabular}}}
        \end{small}
        \label{tab:intermediate_domain}
    \end{minipage}
    \hfill
    \begin{minipage}[hbtp]{0.51\hsize}
        \setlength{\belowcaptionskip}{0.1cm}
        \caption{Investigation of %tail cuboid over-sampling with class-balanced lovasz loss~\cite{berman2018lovasz} on 
        pseudo label class imbalance issue on 3D-FRONT $\rightarrow$ ScanNet}
        \begin{small}
        \scalebox{0.85}{
        \setlength{\tabcolsep}{7mm}{
        \begin{tabular}{c|c}
            \bottomrule[1pt]
            Method & mIoU\\
            \hline
            Noisy Student & 48.13\\
            \hline
            TACM & \textbf{51.42}\\
            CM + lovasz loss~\cite{berman2018lovasz} & 51.68 \\
            TACM + lovasz loss~\cite{berman2018lovasz} & 52.50 \\
            \toprule[0.8pt]
        \end{tabular}}}
        \end{small}
        \label{tab:tail-class}
    \end{minipage}
    }}
\end{table}

\section{Limitations and Open Problems}
Although our model largely closes the domain gaps across simulation and real-world datasets, we still suffer from the inherent limitations of the simulation data. For some categories such as beam and column, the simulator fails to generate realistic shape patterns, resulting in huge negative transfer. Besides, room layouts need to be developed by experts, which may limit the diversity and complexity of the created scenes. Therefore, in order to make simulation data benefit real-world applications, there are still several open problems: how to handle the failure modes of the simulator, how to unify the adaptation and simulation stage in one pipeline, and how to automate the simulation process, to name a few.

%%%%%%%%%%%%%%%%%%%%%%%%
% CONCLUSION
%%%%%%%%%%%%%%%%%%%%%%%%

\section{Conclusions}
We have presented DODA, a data-oriented domain adaptation method with virtual scan simulation and tail-aware cuboid mixing for 3D indoor sim-to-real unsupervised domain adaptation.
Virtual scan simulation generates a more transferable model by mitigating the real-and-simulation point pattern gap. 
Tail-aware cuboid mixing rectifies context biases through creating a tail-aware intermediate domain and facilitating self-training to effectively leverage pseudo labeled target data, further reducing domain gaps.
Our extensive experiments not only show the prominent performance of our DODA in two sim-to-real UDA tasks, but also illustrate the potential ability of TACM to solve general 3D UDA scene parsing tasks. More importantly, we have built the first benchmark for 3D indoor scene unsupervised domain adaptation, including sim to real adaptation and cross-site real-world adaptation. The benchmark suit will be publicly available. We hope our work could inspire further investigations on this problem.
\\[12pt]
\noindent\textbf{Acknowledgement.} This work has been supported by Hong Kong Research Grant Council - Early Career Scheme (Grant No. 27209621), HKU Startup Fund, and HKU Seed Fund for Basic Research.

\bibliographystyle{splncs04}
\bibliography{egbib}

% ===== supplementary materials =====
\clearpage
% \newpage
\appendix

\renewcommand{\thesection}{S\arabic{section}}
\renewcommand{\thetable}{S\arabic{table}}
\renewcommand{\thefigure}{S\arabic{figure}}

\centerline{\large{\textbf{Outline}}}
\vspace{0.5cm} 
\noindent{This supplementary document is arranged as follows:}
\begin{itemize}
    \item[$\bullet$] Sec.~\ref{sec:vss} elaborates the visible range design in occlusion simulation of VSS;
    \item[$\bullet$] Sec.~\ref{sec:tacm} illustrates visualization and analysis of TACM and other data-mixing methods;
    \item[$\bullet$] Sec.~\ref{sec:benchmark} presents the implementation details of benchmark setup for sim-to-real settings and cross-site settings;
    \item[$\bullet$] Sec.~\ref{sec:tcos} presents the per-class results of tail cuboid over-sampling in TACM;
    \item[$\bullet$] Sec.~\ref{sec:cross-site} benchmarks DODA with other popular UDA methods on cross-site settings;
    \item[$\bullet$] Sec.~\ref{sec:nyu} investigates DODA performance on 3D-FRONT $\rightarrow$ NYU-Depth V2, which focuses on the adaptation from simulation 3D to real RGBD;
    \item[$\bullet$] Sec.~\ref{sec:pseudo} analyzes the pseudo-label quality with VSS and TACM.
    \item[$\bullet$] Sec.~\ref{sec:vis} presents the qualitative results of S3DIS and ScanNet on sim-to-real settings.
\end{itemize}

% =====================================
% =====================================
\section{Visible Range Design}\label{sec:vss}
In this section, we elaborate the visible range design. Given the camera position $v$ and the point of interest $h$, the maximum visible range $R_v$ is determined by FOV configurations encompassing the horizontal viewing angle $\alpha_h$, the vertical viewing angle $\alpha_v$ and the viewing mode $\eta$. Specifically, the horizontal visible range $R_v[xy]$ is determined by $\alpha_h$ as Eq.~\eqref{eq:fov_xy}:
\begin{equation}\label{eq:fov_xy}
    R_v[xy] = \left\{p \mid \frac{(p_{xy} - v_{xy})^T(h_{xy} - v_{xy})}{||p_{xy} - v_{xy}||_2||h_{xy} - v_{xy}||_2} > \cos \frac{\alpha_h}{2} \right\},
\end{equation}
where the subscript $xy$ stands for the coordinate vector projected onto the X-Y plane.
\begin{figure}[htbp]
	\centering
	\includegraphics[width=\linewidth]{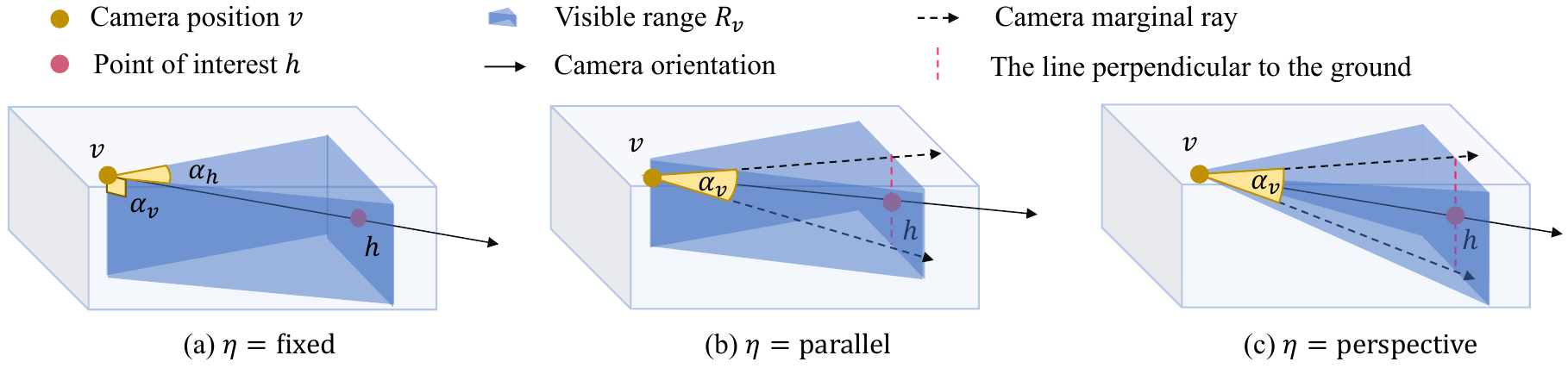}
	\caption{An illustration of visible range with different viewing modes $\eta$. Note that for three modes, the definition of $\alpha_h$ is the same thus we only show it in the fixed mode.}
    \label{fig:vss_2}
\end{figure}
As for the vertical visible range $R_v[z]$, it depends on $\alpha_v$ and $\eta$ as shown in Fig.~\ref{fig:vss_2}. Specifically, for the simplest fixed mode ($\eta=$ fixed), it selects the visible range lower than the horizontal plane passing through camera $v$ if the camera look downwards (see Fig.~\ref{fig:vss_2} (a)); otherwise range above the horizontal plane through $v$ will be selected. In this regard, $\alpha_v$ is fixed at 90$\degree$. 
More flexibly, the parallel mode ($\eta=$parallel) decides the upper and lower bound of vertical visible range as the intersections of marginal rays and the line through $h$ perpendicular to the ground (See Fig.~\ref{fig:vss_2} (b)).
The perspective mode ($\eta=$perspective) further constrains the visible range into a rectangular pyramid bounded by the camera marginal rays (see Fig.~\ref{fig:vss_2} (c)), which is the most sophisticated and realistic camera projection process. Formally, the vertical range $R[v]$ with different viewing modes can be expressed as Eq.~\eqref{eq:fov_z}.

\begin{equation}\label{eq:fov_z}
    \resizebox{.96\hsize}{!}{$
    R_v[z] = 
    \begin{cases}
    \left\{ p \mid p_z > v_z\right\}~\text{if}~h_z>v_z,~\text{otherwise}~ \left\{ p \mid p_z < v_z \right\}, & \eta=\text{fixed}, \\
    \left\{p \mid
    ||v_{xy}-h_{xy}|| \tan{(\theta - \frac{\alpha_v}{2})} < (p_z - v_z)
    < ||v_{xy}-h_{xy}|| \tan{(\theta + \frac{\alpha_v}{2})} \right\}, & \eta=\text{parallel}, \\
    \left\{p \mid
    ||v_{xy}-p_{xy}|| \tan{(\theta - \frac{\alpha_v}{2})} < (p_z -v_z)
    < ||v_{xy}-p_{xy}|| \tan{(\theta + \frac{\alpha_v}{2})} \right\}, & \eta=\text{perspective}, \\
    \end{cases}
    $}
\end{equation} where $\theta$ is the camera pitch angle defined as $\arcsin(\frac{v_z - h_z}{||v-h||_2})$ and $||\cdot||$ denotes the $L_2$ distance. Finally we obtain the visible range $R_v$ as the intersection of $R_v[xy]$ and $R_v[z]$.

% =====================================
% =====================================
\section{Visualization Comparison and Analysis between TACM and Other Data-mixing Methods}\label{sec:tacm}

\begin{figure}[htbp]
	\centering
	\includegraphics[width=\linewidth]{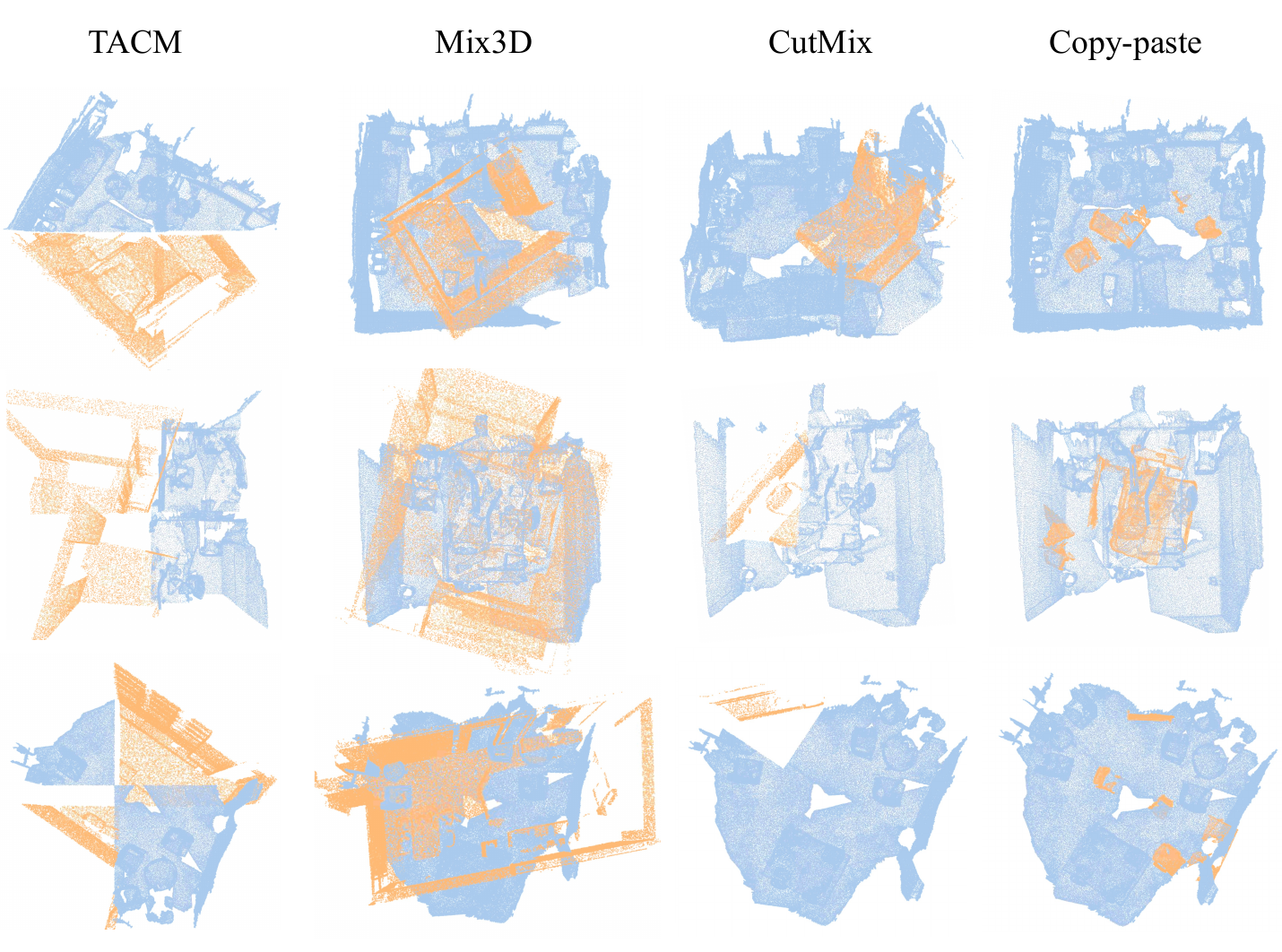}
	\caption{An illustration of TACM examples along with other data-mixing methods. The yellow points are from source scenes and the blue points are from target scenes.}
    \label{fig:tacm_2}
\end{figure}

Even though we already present experimental results in Table 8 in the main paper, to better demonstrate the priority of our TACM among other data-mixing methods, we also show some visualization examples here.
As shown in Fig~\ref{fig:tacm_2},  when scenes are mixed in Mix3D~\cite{nekrasov2021mix3d}, it leads to ambiguity and loss of semantic cues since the neighboring relationship in local areas has been disrupted by mixed points from two domains. As for Copy-paste~\cite{ghiasi2021simple} and CutMix~\cite{yun2019cutmix}, they perturb a local area with randomly sampled patches or instances, which break the local context while introducing no disruptions of the broader context. In contrast, our TACM mixes scenes with the cuboid as the smallest unit, which preserves the local context while also bringing diversity to the global context by different cuboid combinations.

% =====================================
% =====================================
\section{Benchmark Setup}\label{sec:benchmark}

\subsection{Comparison of Large-scale Simulation Datasets.} In our sim-to-real adaptation benchmark, we select 3D-FRONT~\cite{fu20213d} as the source domain which contains 18,968 professionally designed rooms with 13,151 CAD 3D furniture objects from 3D-FUTURE~\cite{fu20213dfuture}. Regarding other large-scale synthetic datasets, SUNCG~\cite{song2017semantic} is not publicly available. Structured3D~\cite{zheng2020structured3d} does not provide interior 3D furniture objects that populate the scenes, which cannot be used as a source dataset without instance classes and layouts. OpenRoom~\cite{li2021openrooms} only contains 2.5K CAD models as the objects, which constrains its diversity. Hence, 3D-FRONT is a favorable choice with adequate scenes as well as professional layouts.

\subsection{Label Mapping.} Due to the different label spaces of datasets, we need to condense common categories for each adaptation task. We manually determine the category mapping relations according to the class names and representative shapes for each class in different datasets.
The selected common classes and mapping relations for 3D-FRONT $\rightarrow$ ScanNet, 3D-FRONT $\rightarrow$ S3DIS, 3D-FRONT $\rightarrow$ NYU-Depth V2 and ScanNet $\leftrightarrows$ S3DIS are shown in Table~\ref{tab:3dfront-scannet}, ~\ref{tab:3dfront-s3dis},~\ref{tab:3dfront-nyu} and~\ref{tab:scannet-s3dis}, respectively. 

\subsection{Implementation Details}
%%%%\vspace{-0.1cm}
\textbf{Network Details.}
We validate DODA on the sparse-convolution-based U-Net backbone~\cite{graham20183d, choy20194d}, which is a popular and  high-performance network on 3D segmentation tasks. The voxel size for point cloud voxelization is set to 2cm.

%%%%\vspace{3pt}
\noindent
\textbf{Training Details.}
In the pretrain stage, we train source data for 11k iterations with 32 batch size on 8 GPUs. SGD optimizer is employed with 0.9 momentum and 0.0001 weight decay. The learning rate is initialized as 0.005 without decay. For pseudo label generation, we set the pseudo label confidence threshold $T$ to 0.7 for ScanNet and to per-class 30\% for S3DIS, to achieve the highest performance. In the self-training stage, we fine-tune the pertrain model for 3.8k iterations on ScanNet and 0.6k iterations on S3DIS. The initial learning rate is set as 0.005 and decayed following the polynomial policy with 0.9 power. The same batch size and optimizer are utilized as in the pretrain stage. The loss trade-off factor $\lambda$ is set as 0.5. During the two stages, commonly used augmentations are applied, in terms of rotation along vertical axis, flip, elastic distortion, jittering and point shuffling. All experiments are conducted on 8 NVIDIA GTX 2080 Ti GPUs.

For the hyper-parameters of VSS, the number of cameras $n_v$ is set to 4 by default. We set the $\alpha_h$ as 180$\degree$, $\alpha_v$ as 90$\degree$ and $\eta$ as fixed for FOV configuration. The point jittering intensity $\delta_p$ is set as 0.01. For cuboid mixing in TACM, the permutation probability $\rho_{s}$ and domain mixing probability $\rho_{m}$ are both set as 0.5. The number of partitions $(n_x,n_y,n_z)$ is set to (2,2,1) with partition perturbations $\delta_\phi$ as 0.1. Thus a total of 4 cuboids are partitioned for each scene. As for tail cuboid over-sampling, we typically set the tail cuboid queue size $N_q$ as 256 and the number of tail classes $n_r$ as 2. The least number of tail cuboids per scene $u$ is set as 2. 

\begin{table*}[htbp]
    \centering
    \caption{Label mapping for 3D-FRONT $\rightarrow$ ScanNet.}
    % \vspace{-0.3cm}
    \begin{small}
      \scalebox{1.0}{
        \setlength{\tabcolsep}{2mm}{
        \begin{tabular}{c|c|l}
            \bottomrule[1pt]
            \textbf{Class} & ScanNet & 3D-FRONT \\
            \hline
            \textbf{wall} & wall & \makecell[l]{wallInner; wallOuter; baseboard; wallTop;\\customizedBackgroundModel; wallbottom; \\customizedFeatureWall; \\extrusionCustomizedBackgroundModel} \\
            \hline
            \textbf{floor} & floor & floor\\
            \hline
            \textbf{cabinet} & cabinet & \makecell[l]{children cabinet;  wardrobe; sideboard/side cabinet/\\console table; wine cabinet; wardrobe; TV stand; \\drawer chest/corner cabinet}\\
            \hline
            \textbf{bed} & bed &  \makecell[l]{king-size bed; bunk bed; bed frame; single bed; kids bed}\\
            \hline
            \textbf{chair} & chair &  \makecell[l]{dining chair; lounge chair/cafe chair/office chair;\\ dressing chair; classic Chinese chair; barstool}\\
            \hline
            \textbf{sofa} & sofa &  \makecell[l]{three-seat/multi-seat sofa; armchair; loveseat sofa; \\L-shapped sofa; lazy sofa; chaise longue sofa}\\
            \hline
            \textbf{table} & table & \makecell[l]{coffee table; round end table; dressing table; \\dining table}\\
            \hline
            \textbf{door} & door & door; pocket\\
            \hline
            \textbf{window} & window & window; baywindow\\
            \hline
            \textbf{bookshelf} & bookshelf & bookcase/jewelry armoire\\
            \hline
            \textbf{desk} & desk & desk\\
            \toprule[1pt]
        \end{tabular}}}
    \end{small}
    \label{tab:3dfront-scannet}
\end{table*}

\begin{table*}[htbp]
    \centering
    \caption{Label mapping for 3D-FRONT $\rightarrow$ S3DIS.}
    % \vspace{-0.3cm}
    \begin{small}
      \scalebox{1.0}{
        \setlength{\tabcolsep}{2mm}{
        \begin{tabular}{c|c|l}
            \bottomrule[1pt]
            \textbf{Class} & S3DIS & 3D-FRONT \\
            \hline
            \textbf{wall} & wall & \makecell[l]{wallInner; wallOuter; baseboard; wallTop;\\customizedBackgroundModel; wallBottom; \\customizedFeatureWall; \\extrusionCustomizedBackgroundModel} \\
            \hline
            \textbf{floor} & floor & floor\\
            \hline
            \textbf{chair} & chair &  \makecell[l]{dining chair; lounge chair/cafe chair/office chair;\\ dressing chair; classic Chinese chair; barstool}\\
            \hline
            \textbf{sofa} & sofa &  \makecell[l]{three-seat/multi-seat sofa; armchair; loveseat sofa; \\L-shapped sofa; lazy sofa; chaise longue sofa}\\
            \hline
            \textbf{table} & table & \makecell[l]{coffee table; round end table; dressing table; \\dining table; desk} \\
            \hline
            \textbf{door} & door & door; pocket\\
            \hline
            \textbf{window} & window & window; baywindow\\
            \hline
            \textbf{bookcase} & bookshelf & bookcase/jewelry armoire\\
            \hline
            \textbf{ceiling} & ceiling & \makecell[l]{customizedCeiling; smartCustomizedCeiling; ceiling;\\ extrusionCustomizedCeilingModel}\\
            \hline
            \textbf{beam} & beam & beam\\
            \hline
            \textbf{column} & column & column\\
            \toprule[1pt]
        \end{tabular}}}
    \end{small}
    \label{tab:3dfront-s3dis}
\end{table*}

\begin{table*}[htbp]
    \centering
    \caption{Label mapping for 3D-FRONT $\rightarrow$ NYU-Depth V2.}
    % \vspace{-0.3cm}
    \begin{small}
      \scalebox{1.0}{
        \setlength{\tabcolsep}{1.5mm}{
        \begin{tabular}{c|c|l}
            \bottomrule[1pt]
            \textbf{Class} & NYU-Depth V2 & 3D-FRONT \\
            \hline
            \textbf{wall} & wall & \makecell[l]{wallInner; wallOuter; baseboard; wallTop;\\customizedBackgroundModel; wallBottom; \\customizedFeatureWall; \\extrusionCustomizedBackgroundModel} \\
            \hline
            \textbf{floor} & floor & floor\\
            \hline
            \textbf{cabinet} & cabinet & \makecell[l]{children cabinet;  wardrobe; sideboard/side cabinet/\\console table; wine cabinet; wardrobe; TV stand; \\drawer chest/corner cabinet}\\
            \hline
            \textbf{bed} & bed &  \makecell[l]{king-size bed; bunk bed; bed frame; single bed;\\ kids bed}\\
            \hline
            \textbf{chair} & chair &  \makecell[l]{dining chair; lounge chair/cafe chair/office chair;\\ dressing chair; classic Chinese chair; barstool}\\
            \hline
            \textbf{sofa} & sofa &  \makecell[l]{three-seat/multi-seat sofa; armchair; loveseat sofa; \\L-shapped sofa; lazy sofa; chaise longue sofa}\\
            \hline
            \textbf{table} & table & \makecell[l]{coffee table; round end table; dressing table; \\dining table}\\
            \hline
            \textbf{door} & door & door; pocket\\
            \hline
            \textbf{window} & window & window; baywindow\\
            \hline
            \textbf{bookshelf} & bookshelf & bookcase/jewelry armoire\\
            \hline
            \textbf{desk} & desk & desk\\
            \hline
            \textbf{ceiling} & ceiling & \makecell[l]{customizedCeiling; smartCustomizedCeiling; \\ceiling; extrusionCustomizedCeilingModel}\\
            \toprule[1pt]
        \end{tabular}}}
    \end{small}
    \label{tab:3dfront-nyu}
\end{table*}

\begin{table*}[htbp]
    \centering
    \caption{Label mapping for ScanNet $\rightarrow$ S3DIS and S3DIS $\rightarrow$ ScanNet.}
    % \vspace{-0.3cm}
    \begin{small}
      \scalebox{1.0}{
        \setlength{\tabcolsep}{2mm}{
        \begin{tabular}{c|c|l}
            \bottomrule[1pt]
            \textbf{Class} & ScanNet & S3DIS \\
            \hline
            \textbf{wall} & wall & wall\\
            \hline
            \textbf{floor} & floor & floor\\
            \hline
            \textbf{chair} & chair & chair\\
            \hline
            \textbf{sofa} & sofa & sofa\\
            \hline
            \textbf{table} & table & table \\
            \hline
            \textbf{door} & door & door\\
            \hline
            \textbf{window} & window & window\\
            \hline
            \textbf{bookshelf} & bookshelf & bookcase\\
            \toprule[1pt]
        \end{tabular}}}
    \end{small}
    \label{tab:scannet-s3dis}
\end{table*}

\subsection{UDA Baselines.}
We reproduce 7 popular 2D UDA methods and 1 3D outdoor UDA method as our baselines, encompassing MCD~\cite{saito2018maximum}, AdaptSegNet~\cite{tsai2018learning}, CBST~\cite{zou2018unsupervised}, MinEnt~\cite{vu2019advent}, AdvEnt~\cite{vu2019advent}, Noisy Student~\cite{xie2020self}  APO-DA~\cite{yang2020adversarial} and SqueezeSegV2~\cite{wu2019squeezesegv2}. 
Similar to DODA, for each baseline, we adopt a sparse-convolution-based U-Net backbone~\cite{graham20183d,choy20194d} and a linear fully-connected point-wise classification head as the overall segmentation network. Besides, some modifications are made for adapting to the 3D vision task as below.
For MCD, the U-Net is used as the generator and the point-wise classification head is used as two-branch classifiers. For AdaptSegNet, we employ its single-level adversarial training performed on the output space. Since the output of the segmentation network is the point-wise predictions, we implement the discriminator as a PointNet-like neural network with 3-layer shared MLP and point random downsampling. For MinEnt, we perform point-wise entropy minimization on target data. For AdvEnt, the same discriminator is utilized as in AdaptSegNet to discriminate outputs from different domains. For APO-DA, we also use the UNet as the generator and only attack the linear classification head to generate point-wise adversarial features. As for the self-training pipeline including CBST and Noisy Student, no other modifications are needed. For the 3D baseline SqueezeSegV2, without official implementations, we self-implement the geodesic alignment and domain calibration modules for our indoor UDA task. The intensity rendering module is discarded since it is specified for outdoor data.
% =====================================
% =====================================
\section{Per-class Results of Tail Cuboid Over-sampling}\label{sec:tcos}
We present per-class results of Tail Cuboid Over-Sampling (TCOS) on 3D-FRONT $\rightarrow$ ScanNet in Table~\ref{tab:scannet-tail} to demonstrate that the performance gain mainly comes from boosting tail categories. From target pseudo label statistics, the tail classes for this setting are bookshelf and desk with sampling ratios around 25\% and 75\%, respectively. For desk, the significant improvements around 6\% verifies the effectiveness of our method in addressing the long-tail issue in pseudo labels. 
%  \ry{Do we need to add lovasz loss results here to compare with}

\begin{table*}[htbp]
    \centering
    \caption{Supplementary adaptation results of 3D-FRONT $\rightarrow$ ScanNet in terms of mIoU (\%). We indicate the best adaptation results in \textbf{bold}. $\dagger$ denotes DODA results without tail cuboid over-sampling.}
    % \vspace{-0.3cm}
    \begin{small}
      \scalebox{0.68}{
        \setlength{\tabcolsep}{2mm}{
        \begin{tabular}{c|c|ccccccccccc}
            \bottomrule[1pt]
            Method & mIoU & wall & floor & cab. & bed & chair & sofa & table & door & wind. & bksf. & desk \\
            \hline
            DODA w/o TCOS$^\dagger$ & 50.55 & 72.63 & \textbf{93.98} & \textbf{28.11} & \textbf{65.88} & 71.43 & 53.17 & 57.40 & \textbf{08.53} & 21.76 & \textbf{57.10} & 26.09 \\
            DODA & \textbf{51.42} & \textbf{72.71} & 93.86 & 27.61 & 64.31 & \textbf{71.64} & \textbf{55.30} & \textbf{58.43} & {08.21} & \textbf{24.95} & 56.49 & \textbf{32.06} \\
            \toprule[1pt]
        \end{tabular}}}
    \end{small}
    \label{tab:scannet-tail}
\end{table*}

% =====================================
% =====================================
\section{Experimental Results on Cross-site Adaptation Tasks}\label{sec:cross-site}

In real-to-real cross-site adaptation tasks, scenes collected from different sites or room types suffer considerable domain discrepancies. To verify the effectiveness of TACM in bridging the real-world domain gaps, we compare DODA (only TACM) with other popular UDA methods on ScanNet $\rightarrow$ S3DIS and S3DIS $\rightarrow$ ScanNet in Table~\ref{tab:scannet2s3dis} and Table~\ref{tab:s3dis2scannet}, respectively. Results show that DODA (only TACM) outperforms other methods by a large margin around 6\% $\sim$ 16\% on ScanNet $\rightarrow$ S3DIS and 4\% $\sim$ 18\% on S3DIS $\rightarrow$ ScanNet. It verifies that our TACM can serve as a general module to eliminate source context bias through target cuboid-level contextual patterns complement.

Besides, to evaluate unsupervised domain adaptation methods, we argue that S3DIS is unsuitable as a source dataset since the per-class results of DODA on real-to-real S3DIS $\rightarrow$ ScanNet are even worse than its counterpart on the sim-to-real 3D-FRONT $\rightarrow$ ScanNet setting (see Table 1 of the main paper).
Although real-to-real adaptation theoretically shows smaller domain gaps than sim-to-real settings, S3DIS is rather simple with a small sample size and limited diversity as its scenes are collected only in three buildings of mainly office and educational use, thus resulting in poor performance of adaptation. It illustrates the importance of carefully selecting real-world datasets as the source domain.
Simulated datasets, on the other hand, can be a consistently appealing choice as a source domain with arbitrarily large size, diverse samples and free annotations.

\begin{table*}[htbp]
    \centering
    % \captionsetup{font={small}}
    \caption{\small Adaptation results of ScanNet $\rightarrow$ S3DIS in terms of mIoU (\%). We indicate the best adaptation result in \textbf{bold}. $\dagger$ denotes the self-training results with TACM based on CBST.}
    % \vspace{-0.3cm}
    \begin{small}
      \scalebox{0.8}{
        \setlength{\tabcolsep}{2mm}{
        \begin{tabular}{c|c|cccccccccccc}
            \bottomrule[1pt]
            Method & mIoU & wall & floor & chair & sofa & table & door & wind. & bkcase.\\
            \hline
            Source Only & 54.09 & 64.38 & 94.39 & 76.15 & 25.46 & 70.55 & 28.98 & 28.52 & \textbf{44.31}\\
            \hline
            MCD~\cite{saito2018maximum} & 49.83 & 61.38 & 95.47 & 73.51 & 32.04 & \textbf{75.24} & 36.95 & 08.01 & 16.02 \\
            AdaptSegNet~\cite{tsai2018learning} & 50.28 & 67.75 & 94.47 & 69.13 & 24.77 & 67.71 & 36.32 & 13.54 & 28.57 \\
            CBST~\cite{zou2018unsupervised} & 60.13 & 68.66 & \textbf{96.02} & 84.61 & 55.04 & 63.80 & 33.47 & 35.61 & 43.84 \\
            MinEnt~\cite{vu2019advent} & 55.31 & 71.31 & 94.70 & 68.10 & 39.86 & 68.23 & 35.98 & 22.03 & 42.24 \\
            AdvEnt~\cite{vu2019advent} & 49.86 & 68.83 & 93.87 & 67.37 & 20.77 & 68.11 & 32.67 & 13.74 & 33.50\\
            Noisy student~\cite{xie2020self} & 58.82 & 66.76 & 95.84 & 83.56 & 52.05 & 64.39 & 36.36 & 37.51 & 34.08 \\
            APO-DA~\cite{yang2020adversarial} & 53.47 & 68.70 & 95.62 & 76.69 & 43.01 & 70.53 & 26.22 & 11.63 & 35.37 \\
            \hline
            DODA (only TACM)$^\dagger$ & \textbf{66.52} & \textbf{73.81} & 95.94& \textbf{85.82} & \textbf{70.71} & 64.64& \textbf{42.93} & \textbf{48.25} & 42.09\\
            \hline
            Oracle & 72.51 & 84.89 & 97.63 & 83.72 & 55.26 & 81.47 & 53.94 & 44.61 & 78.55\\
            \toprule[1pt]
        \end{tabular}}}
    \end{small}
    \label{tab:scannet2s3dis}
\end{table*}

% scannet -> s3dis
\begin{table*}[htbp]
    \centering
    \caption{Adaptation results of S3DIS $\rightarrow$ ScanNet in terms of mIoU (\%). We indicate the best adaptation result in \textbf{bold}. $\dagger$ denotes the self-training results with TACM based on Noisy Student.}
    % \vspace{-0.3cm}
    \begin{small}
      \scalebox{0.8}{
        \setlength{\tabcolsep}{2mm}{
        \begin{tabular}{c|c|cccccccccccc}
            \bottomrule[1pt]
            Method & mIoU & wall & floor & chair & sofa & table & door & wind. & bksf.\\
            \hline
            Source Only & 33.43 & 37.87 & 84.01 & 55.26 & 18.32 & 36.15 & 11.43 & 08.58 & 15.81\\
            \hline
            MCD~\cite{saito2018maximum} & 30.65 & 39.50 & 92.76 & 43.74 & 00.00 & 40.57 & 09.67 & 06.03 & 12.88 \\
            AdaptSegNet~\cite{tsai2018learning} & 36.14 & 58.48 & 91.61 & 35.47 & 21.35 & 44.23 & 07.18 & 09.17 & 21.62 \\
            CBST~\cite{zou2018unsupervised} &  43.08 & 45.43 & 90.11 & 67.53 & 35.48 & 56.51 & \textbf{16.94} & 09.65 & 22.97\\
            MinEnt~\cite{vu2019advent} & 39.40 & 58.11 & 90.31 & 51.18 & 24.86 & 44.20 & 08.10 & 10.27 & 28.19\\
            AdvEnt~\cite{vu2019advent} & 38.09 & 58.83 & 90.24 & 41.73 & 28.96 & 40.68 & 10.58 & 08.11 & 25.59 \\
            Noisy student +~\cite{xie2020self} & 44.81 & 55.61 & 92.75 & 65.72 & 37.77 & 57.77 & 12.54 & \textbf{15.25} & 21.09\\
            APO-DA~\cite{yang2020adversarial} & 38.67 & 63.85 & 90.18 & 49.86 & 22.34 & 41.89 & 06.44 & 04.64 & \textbf{30.15} \\
            \hline
            DODA (only TACM) & \textbf{48.47} & \textbf{65.03} & \textbf{94.25} & \textbf{69.23} & \textbf{43.13} & \textbf{58.79} & 03.58 & 13.86 & 29.91\\
            \hline
            Oracle & 80.06 & 86.78 & 96.02 & 89.98 & 84.24 & 82.15 & 51.19 & 64.99 & 85.16\\
            \toprule[1pt]
        \end{tabular}}}
    \end{small}
    \label{tab:s3dis2scannet}
\end{table*}

% =====================================
% =====================================
\section{Experimental Results on Sim 3D $\rightarrow$ Real RGBD task}\label{sec:nyu}

\subsection{Datasets.}
\textbf{NYU-Depth V2}~\cite{Silberman:ECCV12} is a popular RGBD dataset for semantic segmentation. It contains 1,449 densely annotated RGBD images, \ie~795 training samples and 654 validation samples. Each image has a resolution of 640$\times$480, which can be back-projected to a 3D point cloud containing 3077,200 points. It provides 40 semantic categories.

\subsection{Main Results.}
In the main paper, our experiments focus on the sim-to-real adaptation with target scenes reconstructed by RGBD sequences. However, in real-world scenarios, the real scene can be a single RGBD image captured by the depth camera without reconstructions. Therefore, we also investigate the performance of DODA in such a more challenging setting, \ie~ sim 3D $\rightarrow$ real RGBD. As demonstrated in Table~\ref{tab:3dfront-nyu}, DODA significantly outperforms source only by around 14.3\% and improves CBST by around 8.5\%, largely reducing the cross-modal gaps between 3D-FRONT and NYU-Depth V2. Even only equipping source only with VSS, our DODA (only VSS) also shows its superiority, obtaining 6.3\% and 0.6\% gains compared to source only and CBST separately, which demonstrates the effectiveness of VSS in alleviating the point pattern gaps between simulation 3D and real RGBD. Compared to DODA w/o TACM, TACM further enhances the performance by around 3.4\%, largely bridging the context gaps.

% 3dfront -> nyu v2, main results
% \vspace{-0.3cm}
\begin{table*}[htbp]
    \centering
    \caption{Adaptation results of 3D-FRONT~\cite{fu20213d} $\rightarrow$ NYU-Depth V2 in terms of mIoU (\%). We indicate the best adaptation result in \textbf{bold}. $\dagger$ denotes our pretrain generalization results only with VSS.}
    % \vspace{-0.3cm}
    \begin{small}
      \scalebox{1.0}{
        \setlength{\tabcolsep}{2mm}{
        \begin{tabular}{c|c}
            \bottomrule[1pt]
            Method & mIoU \\
            \hline
            % new setting
            % Source Only & 65.41 & 89.09 & 16.95 & 40.07 & 63.28 & 48.63 & 44.39 & 05.31 & 20.27 & 35.52 & 25.42 & 41.30 \\
            Source Only & 17.80\\
            % \hline
            CBST~\cite{zou2018unsupervised} & 23.58 \\
            % \hline
            DODA (only VSS)$^\dagger$ & 24.14 \\
            DODA w/o TACM & 28.74 \\
            DODA & \textbf{32.12} \\
            % \toprule[0.5pt]
            % \bottomrule[0.5pt]
            \hline
            Oracle & 52.88 \\
            \toprule[1pt]
        \end{tabular}}}
    \end{small}
    \label{tab:nyu}
\end{table*}

% =====================================
% =====================================
\section{Analysis of Pseudo label quality}\label{sec:pseudo}
Self-training relies on both pseudo label accuracy and covering ratio (Eq.~\eqref{eq:cover}) for diversity. As shown in Table~\ref{tab:pseudo_label}, DODA (only VSS) generates pseudo labels with around 15.6\% higher mIoU and 7.7\% larger label covering ratio compared to source only, which benefits the follow-up self-training stage.
Besides, TACM also improves the pseudo label quality after the first self-training round by about 3.6\% mIoU and 0.5\% covering ratio, which is supposed to further boost the iterative self-training if applied.
\begin{equation}\label{eq:cover}
    \text{covering ratio} = \frac{\#~\text{pseudo-labeled points}}{\#~\text{all points}}\times 100\%
\end{equation}

\begin{table}[h]
    \vspace{-0.2cm}
    \centering
    \caption{Results of pseudo label quality with threshold $T=0.7$.}
    \begin{small}
    \scalebox{1.0}{
    \setlength{\tabcolsep}{4mm}{
    \begin{tabular}{c|c|c}
        \bottomrule[1pt]
       \multirow{2}{*}{Method} & \multicolumn{2}{c}{pseudo label} \\
            \cline{2-3}
            & mIoU & covering ratio (\%)\\
            \hline
            Source Only & 35.16 & 59.85\\
            DODA (only VSS) & 50.73 & 67.54\\
            \hline
            DODA (w/o TACM) & 53.24 & 81.51\\
            DODA & 56.85 & 82.05\\
            \toprule[0.8pt]
    \end{tabular}}}
    \end{small}
    \label{tab:pseudo_label}
\end{table}

% =====================================
% =====================================
\section{Visualization}\label{sec:vis}
\begin{figure}[h]
	\centering
	\includegraphics[width=\linewidth]{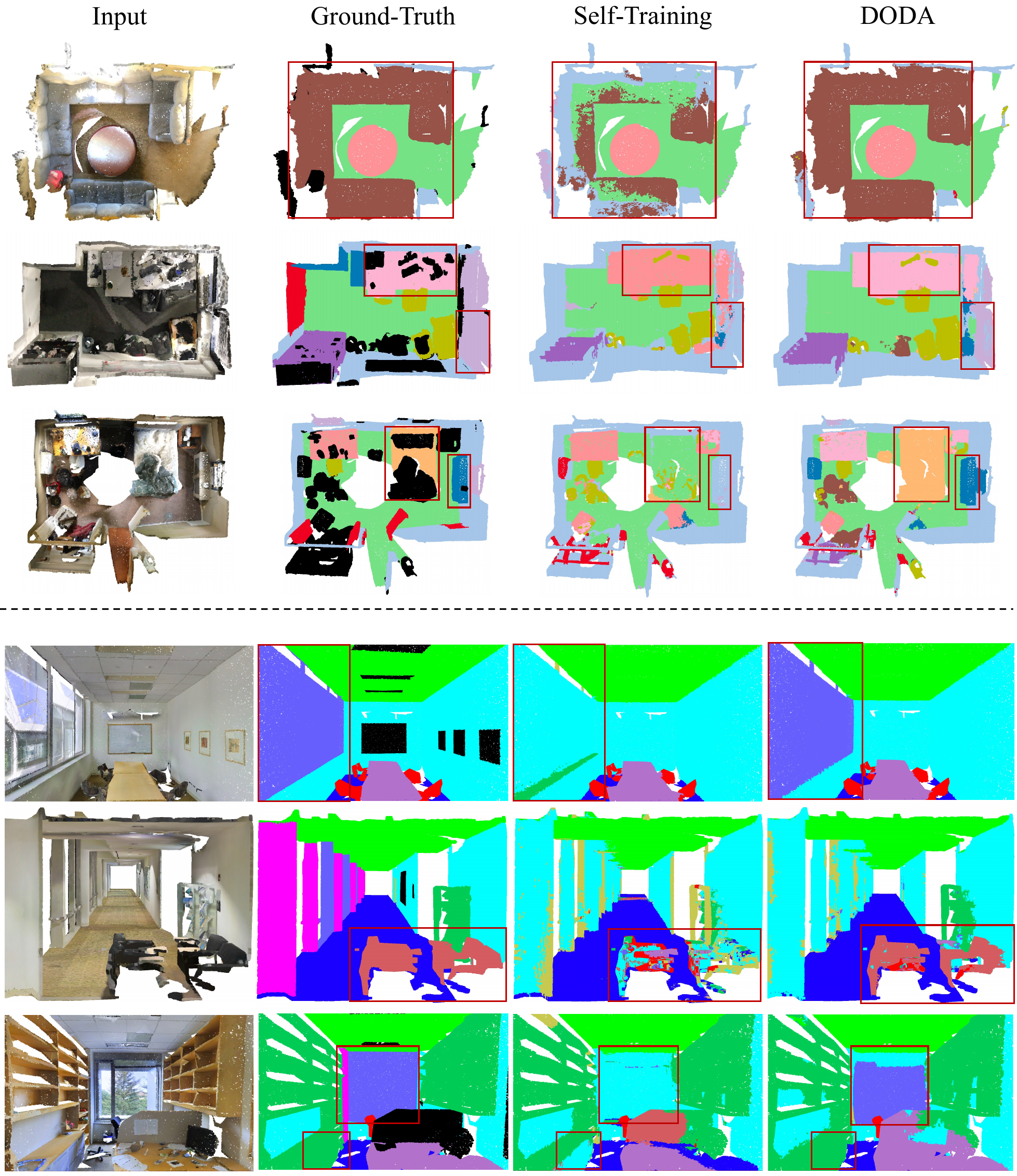}
	\caption{Qualitative results of 3D-FRONT $\rightarrow$ ScanNet (top) and 3D-FRONT $\rightarrow$ S3DIS (bottom). Note that the third column is the prediction of self-training baselines, \ie~Noisy Student for ScanNet and CBST for S3DIS. The red bounding boxes indicate the specific areas where our DODA significantly outperforms self-training baselines.}
    \label{fig:vis}
\end{figure}

We provide some qualitative results of DODA on sim-to-real adaptation tasks of 3D-FRONT $\rightarrow$ ScanNet and 3D-FRONT $\rightarrow$ S3DIS as illustrated in Fig.~\ref{fig:vis}. Compared to self-training baselines, our DODA can segment instances better and generate more accurate and smooth predictions.

\end{document}